\documentclass{article}

\usepackage[preprint]{corl_2026}

\usepackage{times}
\usepackage{multicol}
\hypersetup{
    pdfauthor={Chankyo Kim, Minghan Zhu, Tzu-Yuan Lin, Avantika Rattan, Maani Ghaffari},
    pdftitle={GINIO: A Geometric SO(3)-Equivariant Interface for Neural Inertial Odometry},
    pdfsubject={Accepted CoRL 2026 preprint}
}
\usepackage{tabularx} 
\usepackage{caption}
\usepackage{longtable}
\usepackage{graphicx} 
\usepackage{float} 
\usepackage{mathtools}
\usepackage{amssymb} 
\usepackage{amsmath} 
\usepackage{subcaption}
\usepackage{algorithm, algorithmicx, algpseudocode}
\definecolor{CommentGray}{gray}{0.45}

\usepackage{cleveref} 
\usepackage{booktabs}
\usepackage{balance}
\usepackage{multirow}
\usepackage{arydshln}
\usepackage{siunitx}  
\usepackage{threeparttable}
\usepackage{stfloats}
\usepackage[normalem]{ulem} 
\usepackage{amsthm}
\usepackage{cite}
\usepackage{pifont}
\usepackage{needspace}
\usepackage{wrapfig}

\newtheorem{theorem}{Theorem}

\newtheorem{proposition}[theorem]{Proposition}

\newcommand{\mgj}[1]{}
\newcommand{\tyl}[1]{}

\title{GINIO: A Geometric $\mathrm{SO}(3)$-Equivariant Interface for Neural Inertial Odometry}

\author{
  Chankyo Kim\\
  University of Michigan\\
  Ann Arbor, MI, United States\\
  \texttt{chankyo@umich.edu} \\
  \And
  Minghan Zhu\\
  University of Michigan\\
  Ann Arbor, MI, United States\\
  \texttt{minghanz@umich.edu} \\
  \And
  Tzu-Yuan Lin\\
  University of Michigan\\
  Ann Arbor, MI, United States\\
  \texttt{tzuyuan@umich.edu} \\
  \And
  Avantika Rattan\\
  University of Michigan\\
  Ann Arbor, MI, United States\\
  \And
  Maani Ghaffari\\
  University of Michigan\\
  Ann Arbor, MI, United States\\
  \texttt{maanigj@umich.edu} \\
}

\begin{document}
\maketitle

\begingroup
\renewcommand{\thefootnote}{}
\footnotetext[0]{10th Conference on Robot Learning (CoRL 2026), Austin, Texas, USA.}
\endgroup


\begin{abstract}
Neural inertial odometry increasingly uses networks as learned measurements inside filtering pipelines. Such measurements should transform consistently under arbitrary IMU mounting conventions: their mean must transform as a vector, and their covariance must transform congruently as a second-order tensor. We present \textbf{GINIO}, a geometric \(\mathrm{SO}(3)\)-equivariant interface for neural inertial odometry under arbitrary rotations of the IMU measurement frame. Given calibrated IMU windows, our framework predicts a motion measurement and uncertainty obeying these tensorial laws. To support efficient sensor-frame learning, we introduce Last-Frame Alignment (LFA), a deterministic preprocessing step that is provably equivalent to world-frame training for \(\mathrm{SO}(3)\)-equivariant predictors. The connected estimator tracks sensor-local states such as IMU bias, separating nuisance estimation from the geometric law enforced by the learned measurement. We instantiate the same interface in filter-connected NIO, AirIO-style recurrent aerial prediction, EqNIO-style full-\(\mathrm{SO}(3)\) canonicalization, and ResNet-style temporal backbones. On TLIO, GINIO achieves 2.018 m ID/\(\mathrm{SO}(3)\) ATE while EqNIO degrades to 76.389 m, using \(11.6\times\) fewer FLOPs. On NanoBench, our AirIO-style instantiation improves ATE from 5.579 m to 1.430 m without external attitude input, and our ResNet-style instantiation reaches 0.581 m ATE versus 0.645 m for ResNet1D. On Fetch, GINIO empirically reduces unseen physical-remount ATE from 8.15 m to 0.50 m without retraining, demonstrating robustness beyond the exact coordinate-frame guarantee. For uncertainty, spectral covariance reduces covariance-equivariance error by over three orders of magnitude compared with a diagonal head.

\end{abstract}

\keywords{Equivariant Learning, State Estimation, Learned Inertial Odometry} 


\section{Introduction}

\begin{figure}[t]
\centering
\includegraphics[width=0.95\textwidth]{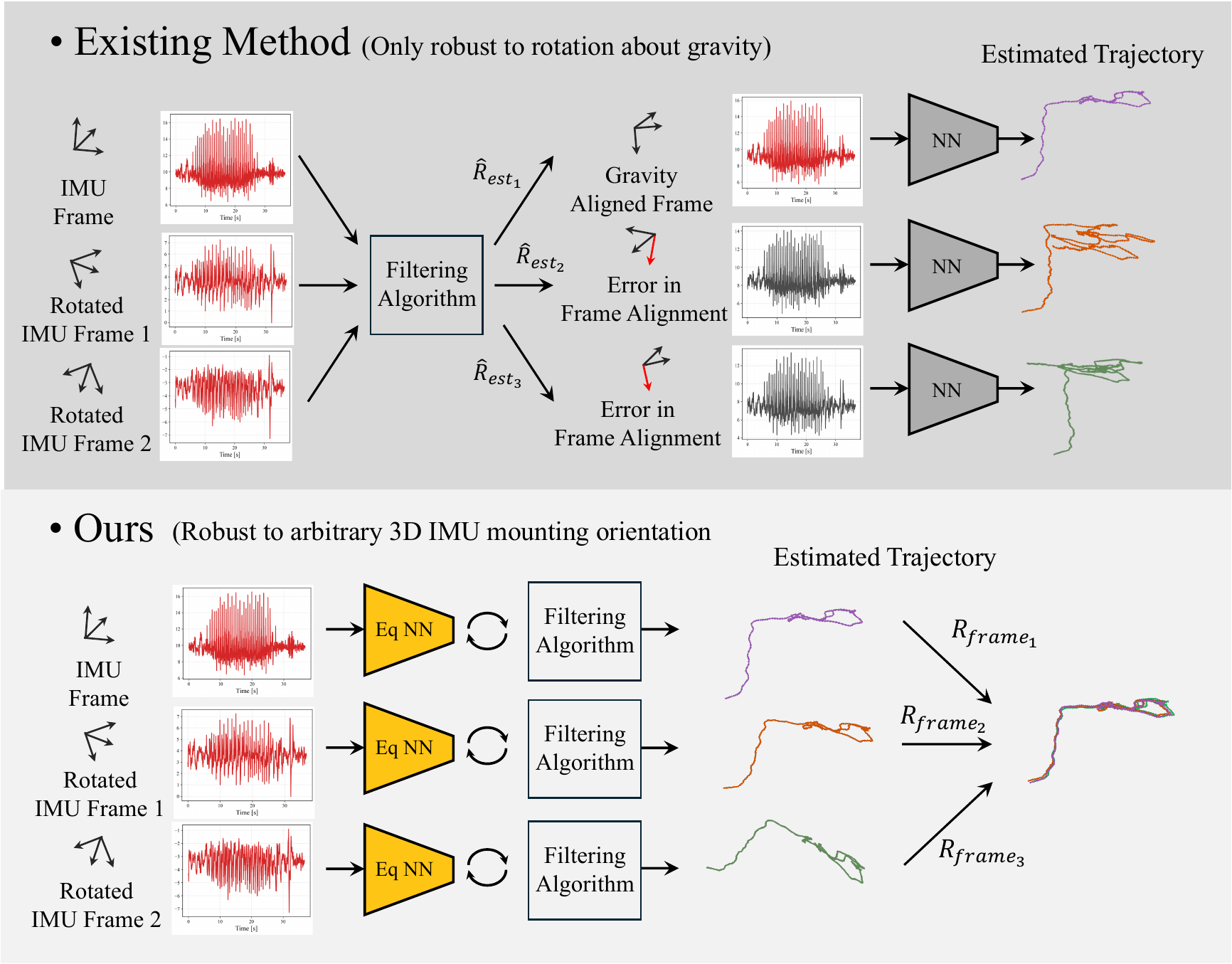}
\caption{\small \(\mathrm{SO}(3)\)-equivariant interface yields consistent learned measurements across mountings.}
\label{fig:equivariance}
\end{figure}

Inertial measurement units (IMUs) enable navigation when vision is unreliable, such as in darkness, rapid motion, visually degraded scenes, or textureless environments~\citep{peng2025aquaticvision}. Classical strapdown integration accumulates bias and noise rapidly, while neural inertial odometry (NIO) reduces drift by learning short-horizon motion priors from IMU windows~\citep{chen2018ionet,yan2018ridi,herath2020RoNIN,liu2020tlio}.  As NIO systems are increasingly used as virtual measurements inside state estimators, the learned measurement must remain meaningful under changes of IMU mounting and sensor-frame convention. The same physical motion can be expressed in different sensor frames when the IMU coordinate convention changes. The coordinate-frame component of an IMU remounting rotates calibrated vector measurements; a physical reinstall may additionally change bias, scale/misalignment, lever-arm effects, vibration, or timing, which are not part of the exact symmetry considered here. Most learned inertial systems reduce mounting variability by rotating measurements into a gravity-aligned frame~\citep{yan2018ridi,herath2020RoNIN,liu2020tlio,jayanth2024eqnio}. This is effective when the gravity frame is accurate, but it introduces a privileged vertical axis and couples the learned input distribution to an external attitude estimate. In dynamic regimes such as aerial motion, body-frame information and rapid attitude changes can be informative rather than nuisance variation~\citep{qiu2025airio}. We instead take a different route: rather than canonicalizing orientation before learning, we impose the rotation law on the learned inertial computation itself.

We use the term \emph{interface} in a restricted sense. GINIO does not prescribe a particular backbone or filter; it specifies how IMU windows are mapped to geometric quantities consumed by an estimator:

\begin{equation}
    \mathcal{G}_{\theta}(\bar{\mathcal{Z}}_{1:n})
    =
    (\hat{\mathbf{m}},\hat{\boldsymbol{\Sigma}}).
\end{equation}
Here \(\hat{\mathbf{m}}\) is a vector-valued inertial measurement and \(\hat{\boldsymbol{\Sigma}}\) is its covariance. Under any rotation \(\mathbf{R}\in\mathrm{SO}(3)\) of the IMU measurement frame, the interface should obey
\begin{equation}
\label{eq:joint_equivariance_intro}
    \mathcal{G}_{\theta}(\rho(\mathbf{R})\bar{\mathcal{Z}})
    =
    \left(
    \mathbf{R}\hat{\mathbf{m}},
    \mathbf{R}\hat{\boldsymbol{\Sigma}}\mathbf{R}^{\mathsf{T}}
    \right).
\end{equation}

Thus motion transforms as a vector and uncertainty transforms as a covariance tensor. Since both are consumed by downstream fusion, both must obey the same geometry.

We propose \textbf{GINIO}, a geometric \(\mathrm{SO}(3)\)-equivariant interface for NIO. Our primary instantiation combines Last-Frame Alignment (LFA), an \(\mathrm{SO}(3)\)-equivariant temporal backbone, spectral covariance, and filter-connected fusion. LFA expresses each IMU window in its final sensor frame and is provably equivalent to world-frame training under full \(\mathrm{SO}(3)\)-equivariance. We further instantiate the same interface in AirIO-style recurrent models, EqNIO-style full-\(\mathrm{SO}(3)\) canonicalized models, and ResNet-style temporal backbones, showing that GINIO is a geometric design principle rather than a backbone-specific trick.

Our contributions are:
\begin{enumerate}
    \item \textbf{A geometric learned interface for NIO} that enforces the \(\mathrm{SO}(3)\) transformation law for both learned motion and covariance.
    \item \textbf{\(\mathrm{SO}(3)\)-equivariant temporal instantiations} across ResNet-style temporal backbones, AirIO-style recurrent models, and full-\(\mathrm{SO}(3)\) EqNIO-style canonicalized models.
    \item \textbf{LFA and tensor-consistent uncertainty}: a sensor-frame preprocessing method equivalent to world-frame training under equivariance, and a spectral covariance head whose uncertainty rotates with the learned motion.
    \item \textbf{Multi-regime validation} across TLIO/RIDI, NanoBench, and physical Fetch remounting, with ablations for augmentation, covariance, capacity, and frame-estimation mismatch.
\end{enumerate}

\section{Related Work}
\label{sec:related_work}

\subsection{Learned Inertial Odometry Across Motion Regimes}

Classical strapdown inertial odometry drifts rapidly with low-cost IMUs because bias, noise, and calibration errors accumulate over time~\citep{titterton2004strapdown,solin2018modeling,hartley2020contact, lin2023proprioceptive}. Data-driven methods reduce this drift by learning motion priors from IMU windows. IONet, RIDI, and RoNIN established neural inertial odometry for pedestrian and mobile sensing, while TLIO and AI-IMU reinterpret neural predictions as learned updates or corrections fused by downstream estimators~\citep{chen2018ionet,yan2018ridi,herath2020RoNIN,liu2020tlio,brossard2020ai, brossard2020denoising,buchanan2022deep,cioffi2023learned,sun2021idol,zeinali2024imunet}. Recent work extends learned inertial odometry to visually degraded and aerial regimes~\citep{peng2025aquaticvision,qiu2025airio}. AirIO highlights the importance of body-frame IMU representation and attitude information for UAV motion~\citep{qiu2025airio}, while NanoBench provides real nano-quadrotor flights with raw IMU, Vicon ground truth, onboard estimator outputs, controller signals, and aggressive maneuvers~\citep{nanobench2026}.

\subsection{Rotation Consistency and Equivariant Models}

Most NIO systems reduce orientation sensitivity by expressing IMU data in a gravity-aligned frame~\citep{yan2018ridi,herath2020RoNIN,liu2020tlio}. This removes much of the roll and pitch variation, but introduces a privileged gravity axis and couples the learned input distribution to an external attitude estimate. Augmentation and auxiliary consistency objectives can improve empirical robustness~\citep{cao2022rio}, but do not make the network equivariant by construction. EqNIO is the closest architecture-level rotation-aware counterpart: it enforces \(O_g(3)\cong O(2)\) subequivariance, the gravity-stabilizer subgroup of \(O(3)\), through a learned canonical frame~\citep{jayanth2024eqnio}. This prior is naturally tied to gravity-stabilized inputs and does not enforce full equivariance to arbitrary three-dimensional remounting rotations. GINIO instead imposes full \(\mathrm{SO}(3)\)-equivariance on the learned computation and covariance. It builds on geometric deep learning tools such as group convolutions, vector neurons, tensor-product networks, and Lie-group equivariant architectures~\citep{cohen2016group,deng2021vector,geiger2022e3nn,lin2024lie,kim2025equivariant}, specialized here to temporal IMU odometry and filter-compatible uncertainty.

\section{Problem Setup and Methodology}
\label{sec:method}

Let \(\bar{\mathbf{z}}^s_i=[(\bar{\mathbf{f}}^s_i)^\top,(\bar{\boldsymbol{\omega}}^s_i)^\top]^\top\in\mathbb{R}^6\) denote a bias-corrected IMU sample with specific force and angular velocity expressed in the sensor frame. A coordinate rotation \(\mathbf{R}\in\mathrm{SO}(3)\) acts on the two vector channels by \(\rho(\mathbf{R})=I_2\otimes\mathbf{R}\), applied componentwise to the window \(\bar{\mathcal{Z}}^s=\{\bar{\mathbf{z}}^s_i\}_{i=1}^n\). Thus, when the same calibrated physical signal is re-expressed in two sensor frames \(s_1,s_2\), it satisfies \(\bar{\mathcal{Z}}^{s_2}=\rho({}^{s_2}_{s_1}\mathbf{R})\bar{\mathcal{Z}}^{s_1}\). Our exact equivariance guarantee concerns this coordinate-frame action on calibrated, bias-corrected vector measurements. A physical IMU remounting may additionally induce bias shifts, scale/misalignment changes, lever-arm effects, vibration changes, or timing offsets; these effects lie outside the exact guarantee and must be handled by calibration and/or the connected estimator. Full measurement equations are provided in Appendix~\ref{app:imu_model}.

GINIO maps a window to a learned motion measurement and covariance,
\begin{equation}
\label{eq:ginio_so3_law}
    \mathcal{G}_{\theta}(\bar{\mathcal{Z}})
    =
    (\hat{\mathbf{m}},\hat{\boldsymbol{\Sigma}}),
    \qquad
    \mathcal{G}_{\theta}(\rho(\mathbf{R})\bar{\mathcal{Z}})
    =
    \left(
    \mathbf{R}\hat{\mathbf{m}},
    \mathbf{R}\hat{\boldsymbol{\Sigma}}\mathbf{R}^{\mathsf{T}}
    \right).
\end{equation}
The learned vector \(\mathbf{m}\) denotes the motion quantity consumed by the backend, such as final velocity or window displacement. In sensor/body-frame settings, Eq.~\eqref{eq:ginio_so3_law} applies directly; in world-frame settings, the learned outputs are represented in world frame to match baseline-native protocols.

\paragraph{Equivariant temporal processing.}
The input window is stored as \(\mathbf{X}\in\mathbb{R}^{n\times C_{\mathrm{in}}\times 3}\), with \(C_{\mathrm{in}}=2\) vector channels and a geometric \(\mathbb{R}^3\) dimension. A channel-mixing layer acts only on vector channels,
\begin{equation}
    \mathbf{Y}_{t,c}=\sum_{c'} w_{cc'}\mathbf{X}_{t,c'},
\end{equation}
and an equivariant temporal convolution similarly uses scalar kernels,
\begin{equation}
    \mathbf{Z}_{t,c}=\sum_{\tau,c'} k_{\tau cc'}\mathbf{Y}_{t-\tau,c'}.
\end{equation}
Because \(w_{cc'}\) and \(k_{\tau cc'}\) are scalars, the same learned coefficient acts on all three geometric coordinates; no arbitrary learned \(3\times3\) matrix mixes spatial axes. These maps therefore commute with the common \(\mathrm{SO}(3)\) action. Nonlinearity is introduced using rotation-invariant scalar gates derived from norms/inner products and equivariant vector interactions such as cross products~\citep{deng2021vector,lin2024lie,kim2025equivariant}. Stacking these operations yields the residual temporal blocks used by GINIO-RD/RS. This structural sharing also explains the parameter efficiency; layer specifications and the recurrent/canonicalized variants are given in Appendix~\ref{app:param_efficiency}.

\paragraph{Spectral covariance.}
For uncertainty, GINIO predicts an equivariant frame \(\mathbf V\in\mathrm{SO}(3)\) from vector features and invariant log-standard deviations \(\lambda_i\). We form \(\mathbf A=[a_1,a_2,a_3]\) from three equivariant vectors and obtain \(\mathbf V\) by SVD-based polar projection to \(\mathrm{SO}(3)\) with determinant correction. The projection is left-equivariant whenever \(\mathbf A\) is full rank. To avoid rank-degenerate frames, we use an \(\mathrm{SO}(3)\)-invariant non-degeneracy regularizer on \(\mathbf A^\top\mathbf A\), encouraging the predicted vectors to remain independent without introducing any preferred direction.
We define \(\sigma_i^2=\exp(2\lambda_i)\) and
\begin{equation}
\label{eq:cov_construction}
    \hat{\boldsymbol{\Sigma}}
    =
    \mathbf{V}\operatorname{diag}(\sigma_1^2,\sigma_2^2,\sigma_3^2)\mathbf{V}^{\mathsf{T}}.
\end{equation}
Since the eigenvalues are invariant scalars, input rotation gives
\(\hat{\boldsymbol{\Sigma}}\mapsto
\mathbf R\hat{\boldsymbol{\Sigma}}\mathbf R^{\mathsf T}\).
Diagonal heads are retained as ablations as they are practical in fixed frames but do not satisfy this congruence law.

\paragraph{Last-Frame Alignment.}
Learning from sensor-frame windows can be difficult because the gravity component rotates within the window. LFA expresses each sample in the final sensor frame. Let \({}^{w}_{s}\mathbf{R}_k\) be the rotation from \(\mathcal{S}(t_k)\) to \(\mathcal{W}\), and define
\begin{equation}
\label{eq:lfa_transform}
    \mathbf{R}_{n,k}=({}^{w}_{s}\mathbf{R}_n)^{\mathsf{T}}{}^{w}_{s}\mathbf{R}_k,
    \qquad
    \bar{\mathbf{z}}^{\mathrm{lfa}}_k=\rho(\mathbf{R}_{n,k})\bar{\mathbf{z}}^s_k .
\end{equation}
LFA does not align to the world vertical; it keeps the representation sensor-centric while reducing within-window rotational nonstationarity.

\begin{proposition}[World-frame--LFA equivalence]
\label{prop:equivariance_agnostic}
Let \(\Phi\) be an \(\mathrm{SO}(3)\)-equivariant predictor whose output is expressed in the same frame as its input sequence. If \(\bar{\mathcal{Z}}^{w}=\{\rho({}^{w}_{s}\mathbf{R}_k)\bar{\mathbf{z}}^s_k\}_{k=1}^{n}\), then
\begin{equation}
    \Phi(\bar{\mathcal{Z}}^{w})
    =
    {}^{w}_{s}\mathbf{R}_n
    \Phi(\bar{\mathcal{Z}}^{\mathrm{lfa}}).
\end{equation}
\end{proposition}
The proof follows from \(\bar{\mathcal{Z}}^{\mathrm{lfa}}=\rho(({}^{w}_{s}\mathbf{R}_n)^{\mathsf{T}})\bar{\mathcal{Z}}^w\) and is given in Appendix~\ref{app:lfa_proof}.

At inference time, \({}^{w}_{s}\mathbf{R}_k\) is supplied by the connected attitude/filter estimate, so LFA retains an estimator-side relative-rotation dependency; Sec.~\ref{subsec:ablation_calibration} stress-tests this dependency.

\paragraph{Filter-connected fusion.}
GINIO follows the common learned-IO pattern of using a neural network as a probabilistic filter update. The filter propagates with raw IMU readings and estimates local states such as bias; the network consumes the corresponding bias-corrected window and outputs \((\hat{\mathbf{m}},\hat{\boldsymbol{\Sigma}})\) for correction. We use SCEKF for baseline-native world-frame comparisons~\citep{liu2020tlio,jayanth2024eqnio} and InEKF in the sensor-frame setting~\citep{barrau2016invariant}. This backend choice is a protocol/deployment choice, not a requirement of the equivariant representation. In the no-filter controlled ablation, the same equivariant predictor gives identical ATE for World and Sensor+LFA after deterministic frame conversion (1.828 m for both; Appendix~\ref{app:ablation_full}). World-frame processing is attractive when an accurate absolute attitude is already available, whereas LFA needs only within-window relative rotations and is therefore useful when absolute attitude is unreliable. The full inference loop is in Appendix~\ref{app:inference_loop}.

Figure~\ref{fig:ginio_pipeline} makes the complete filter-connected data flow explicit. The estimator first propagates the raw IMU and maintains nuisance states such as bias; GINIO then operates on the corresponding calibrated, bias-corrected window expressed in Direct/Body, World, or LFA coordinates. The learned interface---rather than the downstream filter---enforces the \(\mathrm{SO}(3)\) transformation law and returns the motion measurement and uncertainty used in the correction step. Additional estimator-loop and predictor-architecture views are provided in Appendix~Fig.~\ref{fig:ginio_system_arch_appendix}.

\begin{figure*}[t]
\centering
\includegraphics[width=0.9\textwidth]{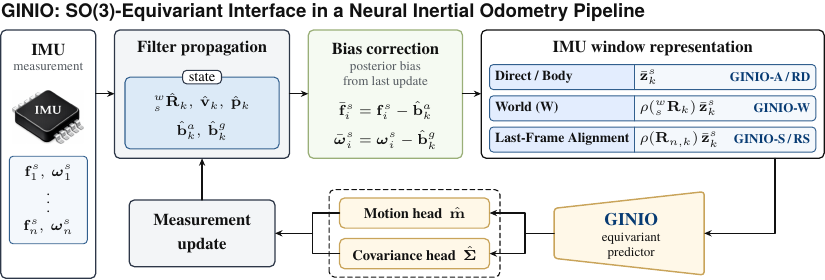}
\caption{Filter-connected GINIO pipeline. The estimator propagates raw IMU measurements and maintains nuisance states such as IMU bias. The learned predictor receives the corresponding calibrated, bias-corrected window in a Direct/Body, World, or Last-Frame-Aligned representation and returns a vector-valued motion measurement together with its covariance for the measurement update. The exact \(\mathrm{SO}(3)\) transformation law is enforced by the learned motion-and-covariance interface, not by the choice of filter backend.}
\label{fig:ginio_pipeline}
\end{figure*}

\section{Experiments}
\label{sec:experiments}

We evaluate GINIO along three axes: filter-connected learned inertial odometry, aerial inertial prediction, and physical IMU remounting. TLIO/RIDI test whether the interface improves standard filter-based NIO; NanoBench tests whether the same \(\mathrm{SO}(3)\) principle transfers to aggressive nano-quadrotor dynamics; and Fetch remounting tests whether the learned predictor remains robust when the IMU is physically reinstalled. Throughout this section, suffixes indicate instantiations rather than separate methods: GINIO-W/S denote world- and sensor-frame variants, GINIO-A denotes the AirIO-style recurrent variant, GINIO-E denotes the full-\(\mathrm{SO}(3)\) EqNIO-style canonicalized variant, and GINIO-RD/RS denote ResNet-style temporal variants with diagonal/spectral covariance heads.

\subsection{Experimental Setup}
\label{subsec:exp_setup}

\paragraph{Datasets.}
TLIO and RIDI serve as human-motion learned-IO benchmarks with standard train/test splits~\citep{liu2020tlio,yan2018ridi}. NanoBench evaluates aerial inertial prediction on real Crazyflie 2.1 flights with Vicon ground truth, raw IMU, onboard estimator outputs, controller signals, and aggressive maneuvers~\citep{nanobench2026}. We further collect a Fetch hardware remounting set in which an IMU is physically reinstalled in unseen orientations. AquaticVision is reported in Appendix~\ref{app:aquatic} as an additional visually degraded evaluation.

\paragraph{Baselines and protocols.}
On TLIO/RIDI, we compare against ResNet+SCEKF baselines (e.g., TLIO), EqNIO, and \(\mathrm{SO}(3)\)-augmented variants in their native protocols~\citep{herath2020RoNIN,liu2020tlio,jayanth2024eqnio}. World-frame rows use SCEKF to match TLIO/EqNIO-style evaluation, while sensor-frame rows use LFA and InEKF. On NanoBench, we compare against AirIO, ResNet1D, and EqNIO O(2), and instantiate GINIO in recurrent, canonicalized, and temporal forms~\citep{qiu2025airio,jayanth2024eqnio}. NanoBench is evaluated as an NN-only aerial prediction benchmark, complementing the filter-connected TLIO/RIDI experiments. ID/ID denotes the standard held-out split; ID/\(\mathrm{SO}(3)\) applies test-time rotations to the IMU measurement frame. Within each benchmark, baselines are evaluated in their native input frame and backend to avoid penalizing method-specific design assumptions. For ID/\(\mathrm{SO}(3)\), we rotate the accelerometer and gyroscope vector channels by a sampled \(\mathbf{R}\in\mathrm{SO}(3)\); vector targets are rotated by the same \(\mathbf{R}\), and covariance targets, when used, are transformed by \(\mathbf{R}\boldsymbol{\Sigma}\mathbf{R}^{\mathsf{T}}\). Predictions are unrotated before evaluation, and all methods are evaluated on the same sampled rotations.

\paragraph{Metrics.}
The main text reports one primary metric per setting: ATE [m] for filter-connected trajectory benchmarks, NanoBench ATE for standard aerial prediction, and velocity RMSE [m/s] for perturbation diagnostics. For the filter-connected human-motion evaluation, RTE is the RMSE of relative translation over fixed \(\Delta t=1\,\mathrm{s}\) windows after removing the yaw error at the start of each window, following the TLIO protocol; the exact expression and drift-ratio definition are given in Appendix~\ref{app:metrics_arxiv}. Additional uncertainty diagnostics, training details, and per-sequence results are provided in the appendix. \textit{To support reproducibility, code, evaluation scripts, and processed benchmark splits will be released upon publication.} 
\subsection{Human-Motion Benchmarks: TLIO and RIDI}
\label{subsec:human_results}

Table~\ref{tab:human_results_compact} summarizes TLIO and RIDI. On TLIO, gravity-stabilized baselines are competitive in the nominal setting but brittle under arbitrary mounting rotations: EqNIO attains the best ID/ID ATE (1.554 m) yet degrades to 76.389 m under ID/\(\mathrm{SO}(3)\). We evaluate EqNIO in its gravity-stabilized protocol with the expected gravity-frame inputs and the same train/test split; the ID/\(\mathrm{SO}(3)\) test rotates the measurement frame. \(\mathrm{SO}(3)\) augmentation improves robustness but remains substantially worse than GINIO. In contrast, GINIO-W remains close to its nominal error while using 0.57M parameters and 17.81M FLOPs, roughly \(10\times\) fewer parameters and \(12\times\) fewer FLOPs than EqNIO. The sensor-frame GINIO-S row confirms that LFA and InEKF provide a deployment path that does not require learned gravity-aligned inputs.

On RIDI, GINIO-W gives the best nominal ATE among world-frame methods, while GINIO-S gives the most stable rotated result in the sensor-frame configuration. Together, TLIO/RIDI show that the same equivariant interface can be used in both baseline-native world-frame filtering and a sensor-frame LFA+InEKF pipeline. Per-sequence nominal ATE distributions for TLIO and NanoBench are provided in Appendix~Fig.~\ref{fig:nominal_cdfs}; they support the intended claim of competitive nominal accuracy rather than uniform nominal superiority.

To broaden the representative NIO baseline coverage without mixing filter-connected and network-only protocols, we also evaluate the learned predictor under the same NN-only RIDI test protocol used for IMUNet~\citep{zeinali2024imunet}. In this matched NN-only comparison, GINIO achieves 0.927 m ATE versus 1.43 m for IMUNet. We report this comparison separately from Table~\ref{tab:human_results_compact} because no downstream filter is involved.

\begin{table}[t]
\centering
\small
\caption{Human-motion benchmarks. World-frame rows follow SCEKF protocols; sensor-frame rows use LFA and InEKF. For RIDI, the ResNet+SCEKF row corresponds to the RoNIN+SCEKF implementation. Bold denotes the best result within each configuration block. Full RTE, drift ratio, and auxiliary metrics are reported in Appendix~\ref{app:full_metrics}.}
\label{tab:human_results_compact}
\setlength{\tabcolsep}{4pt}
\begin{tabular}{lcccccc}
\toprule
\multirow{2}{*}{Method} & \multicolumn{2}{c}{TLIO ATE [m]} & \multicolumn{2}{c}{RIDI ATE [m]} & Params & FLOPs \\
\cmidrule(lr){2-3}\cmidrule(lr){4-5}
& ID/ID & ID/\(\mathrm{SO}(3)\) & ID/ID & ID/\(\mathrm{SO}(3)\) & & \\
\midrule
\multicolumn{7}{l}{\emph{World-frame configuration}} \\
ResNet+SCEKF & 1.725 & 35.699 & 1.207 & 2.990 & 5.42M & 39.44M \\
+ \(\mathrm{SO}(3)\) aug. & 1.844 & 8.687 & 1.401 & 3.011 & 5.42M & 39.44M \\
EqNIO & \textbf{1.554} & 76.389 & 2.408 & 3.460 & 6.02M & 206.78M \\
GINIO-W & 1.615 & \textbf{2.018} & \textbf{0.952} & \textbf{2.450} & 0.57M & 17.81M \\
\midrule
\multicolumn{7}{l}{\emph{Sensor-frame configuration}} \\
ResNet-S + InEKF & 2.814 & 36.593 & 1.445 & 4.752 & 5.42M & 39.44M \\
GINIO-S & \textbf{2.704} & \textbf{2.759} & \textbf{1.319} & \textbf{1.457} & 0.57M & 17.81M \\
\bottomrule
\end{tabular}
\end{table}

\subsection{Aerial Benchmark: NanoBench}
\label{subsec:nanobench}

NanoBench evaluates whether the same geometric interface transfers beyond human motion to aerial dynamics. We use it to separate two questions: whether \(\mathrm{SO}(3)\)-equivariance benefits UAV-specialized recurrent models, and whether equivariant temporal backbones improve aerial prediction. ATE denotes the NanoBench benchmark-aligned trajectory ATE over 17 test sequences. For the AirIO-style recurrent family, we report both no-attitude and with-attitude variants. The no-attitude rows provide the input-matched comparison: neither model receives an external attitude estimate, so the difference isolates the effect of equivariant architecture. Under this fair setting, GINIO-A reduces ATE from 5.579 m to 1.430 m while using \(7.4\times\) fewer parameters than AirIO. The with-attitude rows evaluate the stronger AirIO-native setting in which attitude is provided as an additional cue; GINIO-A still improves ATE from 1.205 m to 1.045 m while using \(8.7\times\) fewer parameters.

In the temporal-backbone family, GINIO-RD achieves 0.581 m ATE, outperforming a 5.43M non-equivariant ResNet1D (0.645 m) and EqNIO O(2) (0.828 m). GINIO-E shows that the same interface can also be instantiated in a full-\(\mathrm{SO}(3)\) canonicalized EqNIO-style model.

Table~\ref{tab:nanobench_compact} also reports ID/\(\mathrm{SO}(3)\) ATE. For non-equivariant baselines, these entries are measured by rerunning inference under rotated test inputs; for GINIO variants, equality follows from the enforced transformation law and is numerically verified. Non-equivariant AirIO and ResNet1D exhibit nonzero rotation gaps, and EqNIO O(2) degrades under full out-of-plane \(\mathrm{SO}(3)\) rotation. In contrast, GINIO-A, GINIO-E, GINIO-RD, and GINIO-RS remain unchanged under the same perturbation. GINIO-RD is the accuracy-oriented diagonal-covariance temporal variant, while GINIO-RS isolates the spectral covariance interface analyzed in Sec.~\ref{subsec:ablation_covariance}; the two rows therefore separate prediction accuracy from uncertainty geometry.

\begin{table*}[t]
\centering
\small
\caption{NanoBench NN-only aerial benchmark with test-time \(\mathrm{SO}(3)\) robustness. ATE follows the benchmark alignment protocol over 17 test sequences. In the AirIO-style block, no-attitude rows provide a comparison without external attitude cues, while with-attitude rows evaluate the AirIO-native attitude-aided setting. GINIO-A denotes the AirIO-style recurrent instantiation; GINIO-E denotes the EqNIO-style full-\(\mathrm{SO}(3)\) canonicalized instantiation; GINIO-RD/GINIO-RS denote ResNet-style temporal instantiations with diagonal/spectral covariance heads.}
\label{tab:nanobench_compact}
\setlength{\tabcolsep}{5pt}
\begin{tabular*}{\textwidth}{@{\extracolsep{\fill}}lccccc}
\toprule
Method & Equiv. & Input & Params & ATE [m] \(\downarrow\) & ID/\(\mathrm{SO}(3)\) ATE [m] \(\downarrow\) \\
\midrule
\multicolumn{6}{l}{\emph{AirIO-style recurrent models}} \\
AirIO (no-att.)        & none  & body      & 0.33M  & 5.579 & 5.679 \\
AirIO (with-att.)      & none  & body+att. & 0.39M  & 1.205 & 1.302 \\
GINIO-A (no-att.)      & SO(3) & body      & 0.045M & 1.430 & 1.430 \\
GINIO-A (with-att.)    & SO(3) & body+att. & 0.045M & \textbf{1.045} & \textbf{1.045} \\
\midrule
\multicolumn{6}{l}{\emph{Temporal and canonicalized models}} \\
ResNet1D               & none  & body            & 5.43M & 0.645 & 0.710 \\
EqNIO O(2)             & O(2)  & gravity-aligned & 6.03M & 0.828 & 1.267 \\
GINIO-E                & SO(3) & gravity-aligned & 6.03M & 0.785 & 0.785 \\
GINIO-RD               & SO(3) & body            & 0.57M & \textbf{0.581} & \textbf{0.581} \\
GINIO-RS               & SO(3) & body+LFA        & 0.80M & 0.789 & 0.789 \\
\bottomrule
\end{tabular*}
\end{table*}

\subsection{Physical IMU Remounting on Fetch}
\label{subsec:fetch}

\begin{table}[t]
\centering
\small
\caption{Physical IMU remounting on Fetch. Models are trained only on the nominal mounting. Unseen remount results average Test Cases 1--2 (46 sequences). Normal uses held-out nominal sequences (4 sequences). Remount nATE denotes ATE normalized by reference path.}
\label{tab:fetch_remounting}
\begin{tabular}{lcccc}
\toprule
Model & Normal ATE [m] & Remount ATE [m] & Remount nATE [\%] & Gain \\
\midrule
ResNet & 0.105 & 8.151 & 226.6 & 1.0\(\times\) \\
GINIO-R & 0.095 & \textbf{0.495} & \textbf{14.0} & \textbf{16.5\(\times\)} \\
\bottomrule
\end{tabular}
\end{table}

\Needspace{0.42\textheight}
\begin{wrapfigure}[15]{r}{0.48\textwidth}
    \centering
    \vspace{0pt}
    \includegraphics[width=\linewidth]{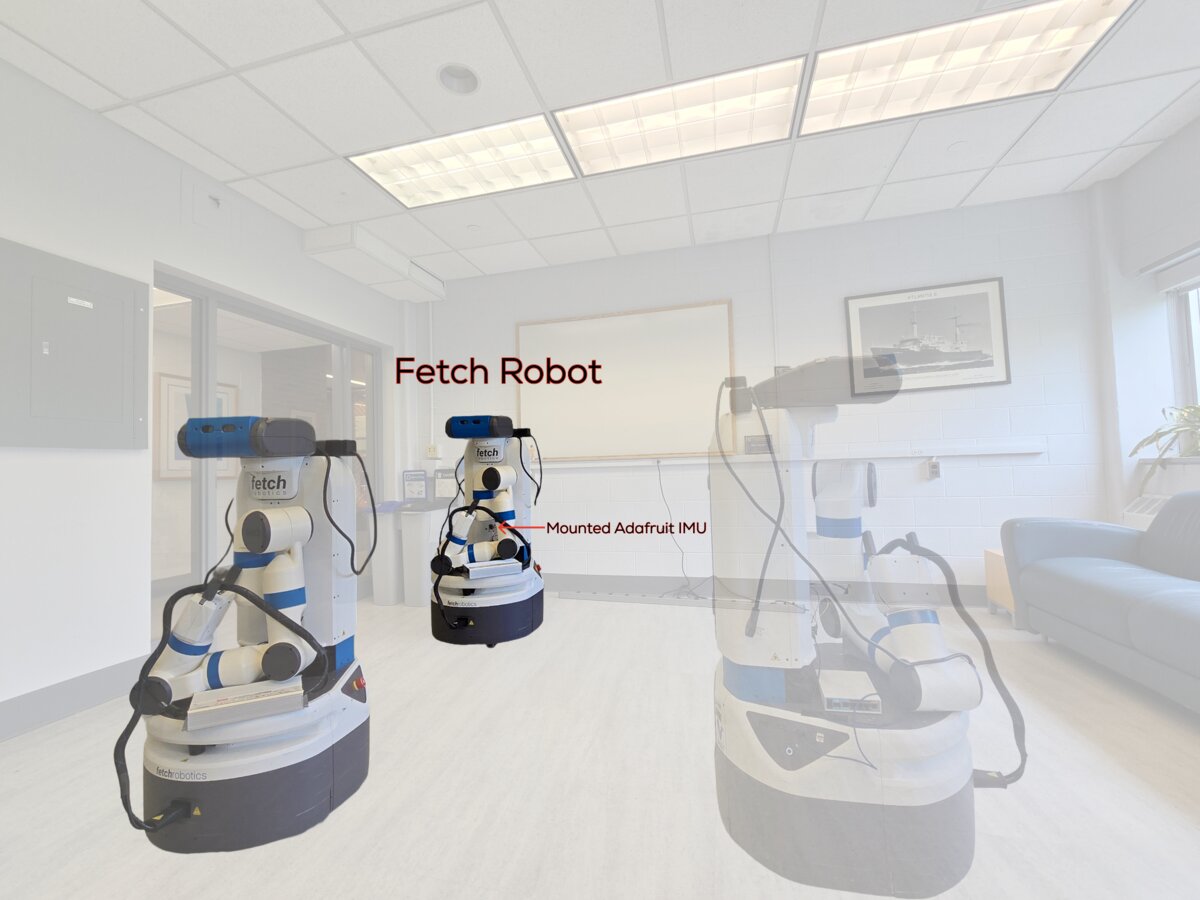}
    \vspace{-0.5em}
    \caption{\small Fetch robot with the IMU mounted inside the mobile base.}
    \label{fig:fetch_robot}
    \vspace{-0.8em}
\end{wrapfigure}
We test physical remounting robustness on a Fetch mobile robot by training only on a nominal IMU mounting and evaluating after reinstalling the IMU in two unseen \(90^\circ\) orientations. The mounted Adafruit BNO055 provides only raw accelerometer and gyroscope channels to the network. The trajectories are manually driven and independently collected per mounting, so the experiment measures unseen mounting generalization rather than paired same-trajectory consistency. We use onboard wheel odometry as the reference and provide full hardware details and representative trajectory diagnostics in Appendix~\ref{app:hardware}.

Table~\ref{tab:fetch_remounting} shows that the non-equivariant ResNet fails under physical remounting, with mean unseen-remount ATE 8.15 m. GINIO-R reduces this to 0.50 m, a \(16.5\times\) improvement, maintaining comparable nominal accuracy. A hardware augmentation ablation in Appendix~\ref{app:hardware_aug} shows that \(\mathrm{SO}(3)\) augmentation narrows the ResNet gap, but does not replace structural equivariance: unaugmented GINIO-R achieves lower unseen-remount ATE while preserving nominal accuracy. Additional \(\mathrm{SO}(3)\) rotations of the recorded IMU streams change GINIO-R ATE by less than \(10^{-3}\) m in Appendix~\ref{app:hardware}.



\section{Ablations and Stress Tests}
\label{sec:ablations}

We isolate the mechanisms behind the main results: architectural equivariance rather than augmentation, tensor-consistent covariance rather than axis-aligned uncertainty, and the boundary between coordinate-frame rotations and frame-estimation mismatch.

\subsection{Architectural Equivariance vs. Augmentation}
\label{subsec:ablation_equivariance}

In our controlled TLIO ablation, architectural equivariance is more reliable than \(\mathrm{SO}(3)\) augmentation for arbitrary 3D rotations. A non-equivariant ResNet fails under ID/\(\mathrm{SO}(3)\) (1.817 m \(\rightarrow\) 108.235 m ATE). \(\mathrm{SO}(3)\) augmentation reduces this error to 3.189 m, but still underperforms the equivariant model, which remains at 1.828 m. The same ablation shows that LFA restores world-frame accuracy in the sensor-frame setting, supporting the WF--LFA equivalence. Matched-backend ablations further show that robustness is not a filter artifact: with the same InEKF backend, a non-equivariant ResNet-S degrades from 2.814 m to 36.593 m under rotation, whereas GINIO-S remains stable at 2.704 m and 2.759 m. Full controlled, capacity, and backend ablations are provided in Appendix~\ref{app:ablation_full}.

GINIO variants are parameter-efficient for structural reasons rather than by pruning or post-hoc compression. Equivariant vector-channel layers share the same learned channel interaction across the geometric \(\mathbb{R}^3\) components, yielding compact temporal backbones: GINIO-RD/RS use 0.57M/0.80M parameters, respectively, versus 5.4M for non-equivariant ResNet-style baselines. The NanoBench recurrent results provide an additional capacity check: GINIO-A uses only 0.045M parameters, \(7.4\times\)--\(8.7\times\) fewer than AirIO, yet improves ATE in both no-attitude and with-attitude settings. This supports the view that equivariance acts as a structural regularizer, especially in data-limited aerial learning. The explicit weight-shape comparison is provided in the supplementary material.

\subsection{Spectral Covariance and Uncertainty Consistency}
\label{subsec:ablation_covariance}

\begin{figure}[t]
\centering
\includegraphics[width=0.9\columnwidth]{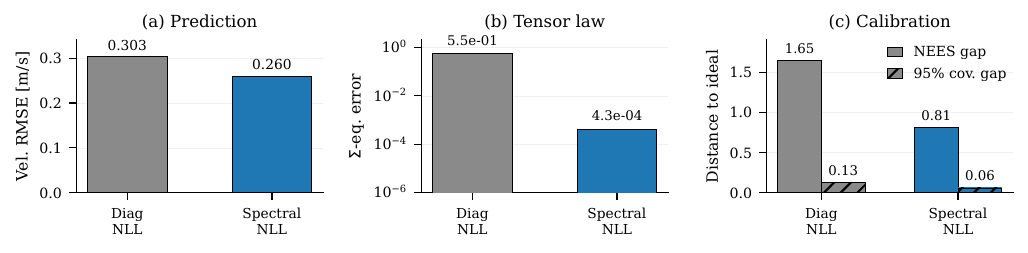}
\caption{NanoBench covariance ablation for NLL-trained heads. Diagonal covariance is axis-aligned and has large tensor-law error. Spectral covariance satisfies the \(\mathrm{SO}(3)\) congruence law and reduces velocity RMSE; calibration is mixed across NEES and coverage metrics rather than uniformly improved. In panel (c), lower is better: NEES gap is \(|\mathrm{NEES}-3|\), and 95\% coverage gap is \(|\mathrm{cov}_{95}-0.95|\).}
\label{fig:nanobench_covariance}
\end{figure}

A vector-valued learned motion update should be paired with a covariance that transforms by congruence. Figure~\ref{fig:nanobench_covariance} compares NLL-trained diagonal and spectral covariance heads on NanoBench. Writing \(\hat{\boldsymbol{\Sigma}}(\bar{\mathcal{Z}})\) for the covariance output of \(\mathcal{G}_{\theta}(\bar{\mathcal{Z}})\), the diagonal head is axis-aligned: it can model uncertainty scale, but it does not generally satisfy the covariance part of Eq.~\eqref{eq:ginio_so3_law},
\begin{equation}
\label{eq:cov_congruence}
\hat{\boldsymbol{\Sigma}}(\rho(\mathbf{R})\bar{\mathcal{Z}})
=
\mathbf{R}\hat{\boldsymbol{\Sigma}}(\bar{\mathcal{Z}})\mathbf{R}^{\mathsf{T}}.
\end{equation}
The spectral head satisfies this tensor law by construction. Empirically, it reduces velocity RMSE and decreases covariance-equivariance error by more than three orders of magnitude. Its nominal calibration is not uniformly better than the diagonal head: the NLL-trained spectral model has NEES closer to the ideal value of 3, while the diagonal head has 95\% coverage closer to the nominal 0.95 target. We therefore treat tensor/geometric consistency, rather than uniform calibration improvement, as the primary benefit of the spectral parameterization. A filter-connected fixed-mean covariance control is shown in Fig.~\ref{fig:filter_connected_sweeps}b, while the full NLL decomposition, coverage curves, and raw covariance metrics are reported in Appendix~\ref{app:cov_details}.

\subsection{Test-Time Frame-Estimation Mismatch}
\label{subsec:ablation_calibration}

Frame rotation and frame-estimation mismatch are related but distinct deployment shifts. A coordinate-frame rotation applies the same \(\mathbf{R}\in\mathrm{SO}(3)\) to the calibrated IMU vectors and to the target motion, and is exactly the symmetry encoded by GINIO. Frame-estimation mismatch instead corrupts the preprocessing frame itself: the gravity frame used by gravity-stabilized pipelines or the relative frame used by LFA. The exact equivariance guarantee does not make such errors disappear, but it changes the failure mode. A predictor that does not bind its learned representation to a privileged gravity axis should degrade more gracefully when the preprocessing frame is imperfect.

\begin{figure*}[t]
\centering
\includegraphics[width=\textwidth]{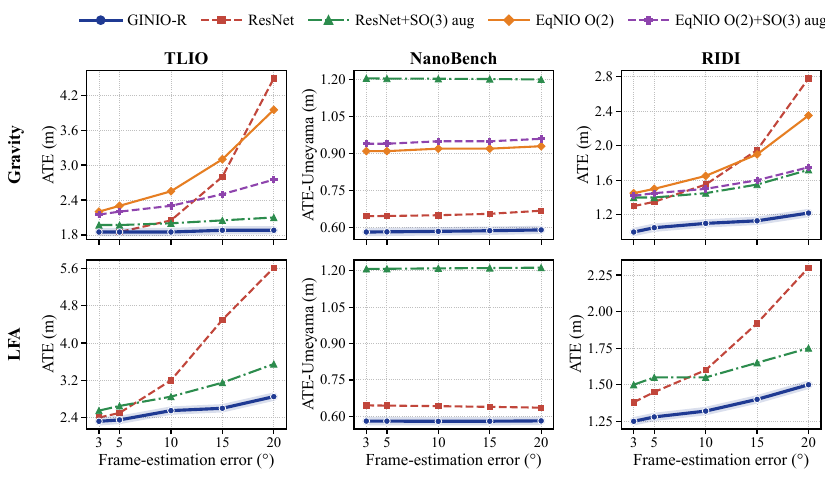}
\caption{Test-time frame-estimation mismatch under gravity-frame and LFA relative-frame errors. The x-axis denotes the injected angular error \(\theta_{\mathrm{err}}\) in the preprocessing frame. Lower ATE is better. Augmented baselines use full \(\mathrm{SO}(3)\) train-time random rotations.}
\label{fig:calib_mismatch_full}
\end{figure*}

We test this by injecting angular error \(\theta_{\mathrm{err}}\) into the preprocessing frame at test time. In the gravity-frame setting, the perturbation corrupts the attitude used by gravity-stabilized pipelines. In the LFA setting, it corrupts the relative rotations used to express each window in the final sensor frame. Figure~\ref{fig:calib_mismatch_full} reports ATE sweeps across TLIO, NanoBench, and RIDI under \(\theta_{\mathrm{err}}\in\{3^\circ,5^\circ,10^\circ,15^\circ,20^\circ\}\).

The result complements the ID/\(\mathrm{SO}(3)\) test. ID/\(\mathrm{SO}(3)\) verifies the exact coordinate-frame symmetry; frame-estimation mismatch tests whether the representation remains stable when the estimated preprocessing frame is imperfect. Across the discriminative TLIO and RIDI settings, GINIO-R shows the flattest or near-flattest degradation under both gravity-frame and LFA relative-frame errors. Full \(\mathrm{SO}(3)\) augmentation helps the non-equivariant baselines, but does not consistently match the stability of an architecturally equivariant predictor. NanoBench is visually near-flat in this stress test, but the measured gravity-frame sweep follows the same trend: GINIO-R has the smallest \(3^\circ\!\rightarrow20^\circ\) ATE increase (\(+0.03\) m), compared with EqNIO O(2) at \(+0.25\) m and its augmented variant at \(+0.10\) m.

\begin{figure*}[t]
\centering
\includegraphics[width=0.98\textwidth]{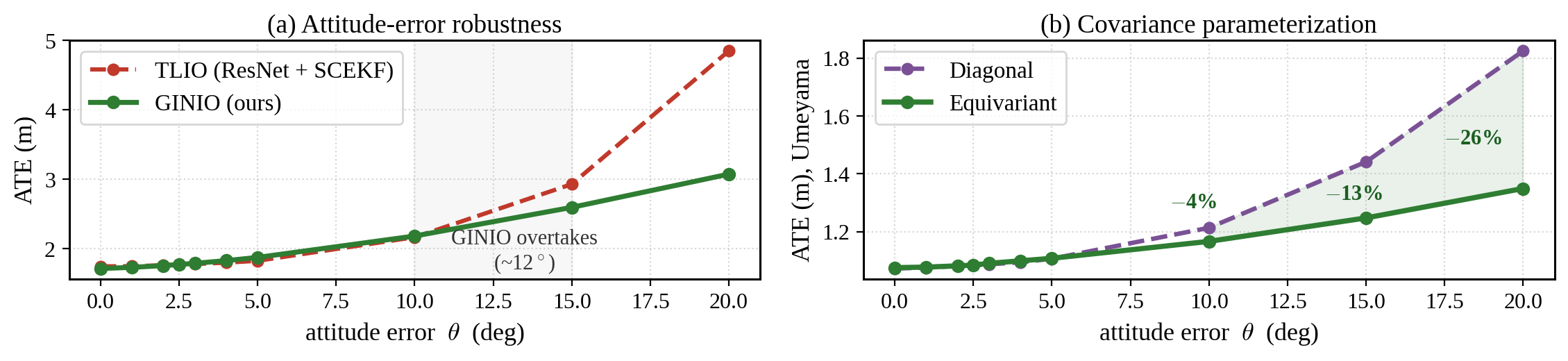}
\caption{Filter-connected TLIO controls with the SCEKF backend held fixed. (a) Attitude/frame-estimation error sweep comparing a TLIO-style ResNet+SCEKF baseline with GINIO. (b) Covariance-only sweep with an identical frozen mean predictor, comparing a coordinate-diagonal uncertainty head with the equivariant spectral-derived parameterization. Holding the backend fixed---and in (b) the learned mean as well---isolates the learned interface and covariance parameterization from filter changes.}
\label{fig:filter_connected_sweeps}
\end{figure*}

To verify that this trend persists once the learned measurement is closed through an estimator, we run two filter-connected TLIO controls with the SCEKF backend held fixed (Fig.~\ref{fig:filter_connected_sweeps}). Panel~(a) compares the TLIO-style non-equivariant ResNet and GINIO under the same injected frame error; at \(20^\circ\), GINIO obtains 3.07 m ATE versus 4.85 m for the baseline. Panel~(b) freezes the same learned mean predictor and varies only the covariance parameterization. The coordinate-diagonal head is marginally better nominally (1.072 vs. 1.081 m), whereas the equivariant spectral-derived covariance is more robust at \(20^\circ\) (1.349 vs. 1.826 m), with paired differences significant for \(\theta\geq15^\circ\). Thus the spectral head is not uniformly superior in the nominal setting; its advantage is geometric consistency and increasing robustness as the frame is perturbed.

These filter-connected controls complement the network-only sweeps rather than replacing them: ID/\(\mathrm{SO}(3)\) tests the exact coordinate-frame law, the frame-estimation sweeps test sensitivity to imperfect preprocessing attitudes, and the closed-loop controls verify that the same robustness trend survives downstream fusion. Representative same-SCEKF trajectories under test-time \(\mathrm{SO}(3)\) frame rotations are shown in Appendix~Fig.~\ref{fig:filter_rotation_trajectories}. Such frame-estimation errors are estimator-side effects whose magnitude depends on the filter estimate, not on the exact coordinate-frame symmetry encoded by the network. The full network-only numeric sweep is provided in Appendix~\ref{app:calibration_mismatch}.

\section{Limitations}
\label{sec:limitations}

GINIO enforces exact equivariance for calibrated, bias-corrected vector measurements under coordinate-frame rotations. It does not eliminate physical-remount or estimator-side effects such as remount-induced bias shifts, sensor scale/misalignment, lever-arm effects, vibration changes, timing offsets, or inaccurate attitude estimates used by preprocessing; these lie outside the exact symmetry guarantee and must be handled by calibration and/or the connected estimator. LFA depends on relative attitude estimates, so severely corrupted relative rotations can degrade the sensor-frame representation; Sec.~\ref{subsec:ablation_calibration} stress-tests this dependency. Our Fetch study uses wheel odometry as reference and independently collected manual trajectories, so it provides empirical evidence of unseen physical-remount robustness beyond the exact coordinate-frame guarantee rather than a paired same-motion repeatability test.

\section{Conclusion}
\label{sec:conclusion}
We presented GINIO, a geometric \(\mathrm{SO}(3)\)-equivariant interface for neural inertial odometry. The central principle is that learned inertial measurements should transform like the physical quantities they represent: motion as a vector and covariance by congruence. Across filter-connected TLIO/RIDI benchmarks, aerial NanoBench prediction, and physical Fetch remounting, the same interface improves robustness to mounting rotations while remaining compatible with different backbones and estimators. These results support \(\mathrm{SO}(3)\) equivariance as a filter-compatible design principle for learned inertial odometry rather than a dataset- or architecture-specific robustness trick.


\clearpage


\bibliography{strings-abrv,ieee-abrv,references}  

\clearpage
\appendix

\section{Supplementary Material Overview}
\label{app:overview}

This appendix provides the details omitted from the main text: the IMU model and inference loop, architecture and parameter-efficiency details, LFA proof, full benchmark metrics, NanoBench protocol details, covariance diagnostics, frame-estimation mismatch sweeps, Fetch hardware validation, and additional auxiliary evaluations.

\section{IMU Model, Group Action, and Inference Loop}
\label{app:imu_model}

\subsection{Full IMU Measurement Model}

We use the standard strapdown IMU model. Let \(\mathcal W\), \(\mathcal B(t)\), and \(\mathcal S(t)\) denote world, body, and sensor frames. The sensor is rigidly attached to the body by a fixed extrinsic rotation \({}^{s}_{b}\mathbf{R}\in\mathrm{SO}(3)\), mapping body-frame vectors into the sensor frame. At time \(t_i\), the gyroscope and accelerometer measure
\begin{equation}
\begin{aligned}
\boldsymbol{\omega}_i^s &= {}^{s}_{b}\mathbf{R}\,\bar{\boldsymbol{\omega}}_i^b + \mathbf{b}_i^g + \boldsymbol{\eta}_i^g,\\
\mathbf{f}_i^s &= {}^{s}_{b}\mathbf{R}\!\left(\bar{\mathbf{a}}_{\mathrm{kin},i}^b - ({}^{w}_{b}\mathbf{R}_i)^\top \mathbf{g}^w\right) + \mathbf{b}_i^a + \boldsymbol{\eta}_i^a,
\end{aligned}
\label{eq:app_imu_model}
\end{equation}
where \(\bar{\boldsymbol{\omega}}_i^b\) and \(\bar{\mathbf{a}}_{\mathrm{kin},i}^b\) are body-frame angular velocity and kinematic acceleration, \(\mathbf{g}^w\) is gravity, and \(\mathbf{b}_i^{(\cdot)}\), \(\boldsymbol{\eta}_i^{(\cdot)}\) are bias and noise. The equivariance statements in the main text are defined on calibrated, bias-corrected vector measurements
\begin{equation}
\bar{\mathbf{z}}_i^s =
\begin{bmatrix}
(\bar{\mathbf{f}}_i^s)^\top & (\bar{\boldsymbol{\omega}}_i^s)^\top
\end{bmatrix}^{\top}
\in\mathbb R^6 .
\end{equation}
The coordinate-frame component of a mounting change is represented by \(\mathbf{R}\in\mathrm{SO}(3)\), acting on the two calibrated vector channels by \(\rho(\mathbf{R})=I_2\otimes \mathbf{R}\). Thus, re-expressing the same calibrated physical signal in a rotated sensor frame maps the IMU window by \(\rho(\mathbf{R})\). Additional physical-remount effects such as bias shifts, scale/misalignment, lever-arm effects, vibration, or timing changes are not part of this exact group action.

\subsection{Filter-Connected Inference Loop}
\label{app:inference_loop}
GINIO is used as a learned probabilistic measurement rather than as a standalone
inertial integrator. The estimator propagates with raw IMU readings and maintains local
nuisance states such as IMU bias. At each update, the network consumes a bias-corrected
window \(\bar{\mathcal Z}_k\) and predicts a motion measurement \(\hat{\mathbf{m}}_k\) with
covariance \(\hat{\boldsymbol{\Sigma}}_k\). Writing \(h(\cdot)\) for the measurement function induced by
the current filter state \(\mathbf{x}_k\), the filter treats the network output as
\begin{equation}
\hat{\mathbf{m}}_k = h(\mathbf{x}_k) + \mathbf{n}_k,
\qquad
\mathbf{n}_k \sim \mathcal N(0,\hat{\boldsymbol{\Sigma}}_k),
\end{equation}
and applies the corresponding Gaussian update. For sensor-frame variants we first apply
Last-Frame Alignment to express the window in the final sensor frame, and rotate
\((\hat{\mathbf{m}}_k,\hat{\boldsymbol{\Sigma}}_k)\) into the backend measurement frame before the update.

We use a stochastic-cloning EKF (SCEKF) for baseline-native world-frame comparisons and
an invariant EKF (InEKF) for the sensor-frame setting. The two are paired with the
frontend by design: because GINIO predicts a quantity that transforms as a vector under
\(\mathrm{SO}(3)\) and a covariance that transforms by congruence
(\(\hat{\boldsymbol{\Sigma}}\mapsto \mathbf{R}\hat{\boldsymbol{\Sigma}}\mathbf{R}^\top\)), its output is naturally expressed in the same
group structure that the InEKF uses to track error. For the InEKF the state lives on a
matrix Lie group \(G\), and we use the left-invariant error
\(\boldsymbol{\eta}_k^L=\hat{\mathbf{X}}_k^{-1}\mathbf{X}_k\), with Lie-algebra coordinates
\(\boldsymbol{\xi}_k=\operatorname{Log}(\boldsymbol{\eta}_k^L)^\vee\). Let \(\operatorname{Exp}_G(\boldsymbol{\delta}) := \exp(\boldsymbol{\delta}^\wedge)\) map vector Lie-algebra coordinates to the group. Consistently, the correction is applied on the right:
\begin{equation}
\hat{\mathbf{X}}_k^{+}=\hat{\mathbf{X}}_k\,\operatorname{Exp}_G\!\big(\mathbf{K}_k\,\mathbf{r}_k\big),
\end{equation}
where \(\mathbf{r}_k\) is the measurement residual and \(\mathbf{K}_k\) the Kalman gain. This invariant error-coordinate form is compatible with changes of the global coordinate frame; it is the filter-side representation choice and does not itself provide the frontend's equivariance. Bias and other local nuisance states remain estimator-side
quantities.

\section{Architecture Details and Parameter Efficiency}
\label{app:architecture_details}

\begin{figure*}[t]
    \centering
    \begin{subfigure}[t]{\textwidth}
        \centering
        \includegraphics[width=\linewidth,trim=74 35 33 23,clip]{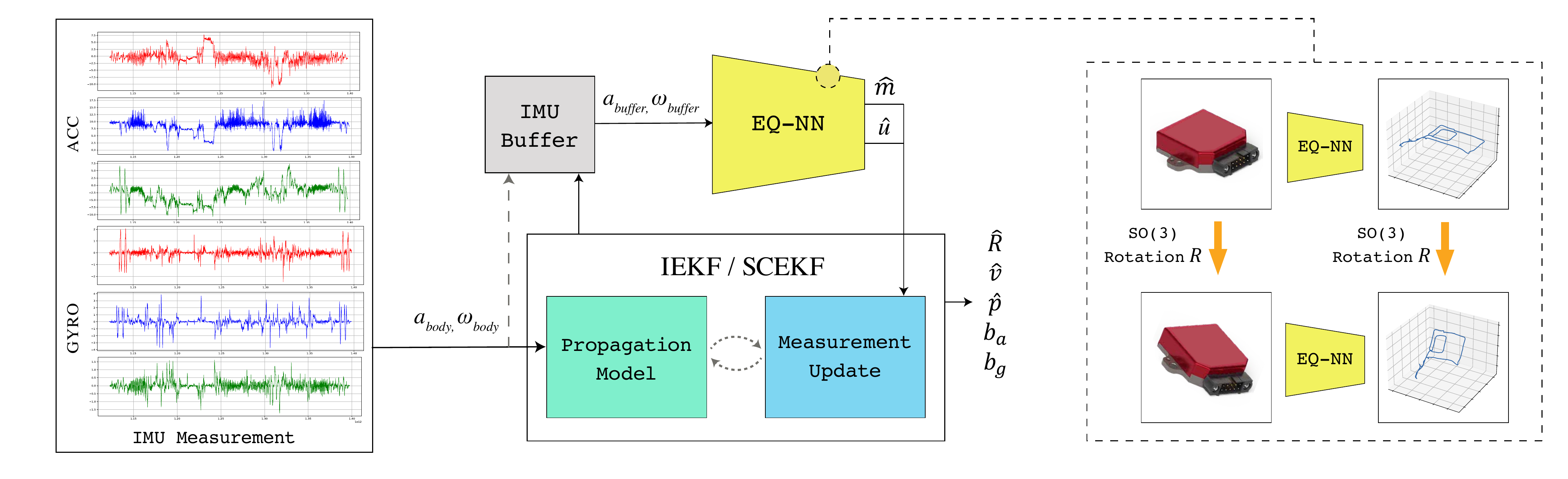}
        \caption{Filter-connected learned-measurement loop. The filter propagates using raw IMU measurements, estimates nuisance states such as bias, and consumes the learned motion and covariance measurement.}
    \end{subfigure}

    \vspace{0.8em}
    \begin{subfigure}[t]{\textwidth}
        \centering
        \includegraphics[width=\linewidth]{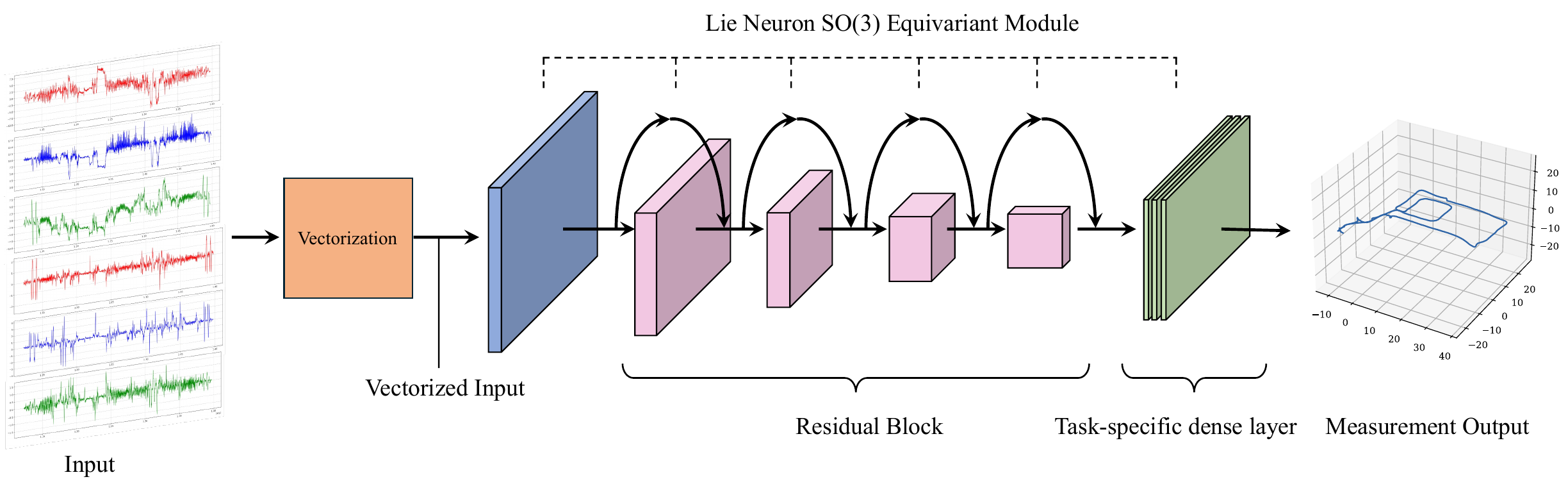}
        \caption{Equivariant temporal predictor overview. Vector-valued features are processed using \(\mathrm{SO}(3)\)-equivariant operations before the task-specific motion and uncertainty heads.}
    \end{subfigure}
    \caption{Additional implementation details of GINIO. (a) Integration of the learned measurement interface with the filter-connected estimator. (b) High-level organization of the equivariant temporal predictor. These views complement the end-to-end pipeline in Fig.~\ref{fig:ginio_pipeline}.}
    \label{fig:ginio_system_arch_appendix}
\end{figure*}

Figure~\ref{fig:ginio_system_arch_appendix} provides complementary implementation views of the estimator integration and equivariant predictor architecture. Unlike the end-to-end data flow in Fig.~\ref{fig:ginio_pipeline}, these diagrams emphasize the learned-measurement loop and the high-level organization of the temporal predictor.

\subsection{Equivariant Temporal Layers}

The input window is stored as \(\mathbf{X}\in\mathbb R^{n\times C_{\rm in}\times 3}\), where the final dimension is the geometric \(\mathbb R^3\) acted on by \(\mathrm{SO}(3)\). Equivariant channel mixing uses scalar weights over vector channels. For vector features \(\mathbf{X}\in\mathbb R^{C_{\rm in}\times 3}\), a channel-mixing layer has the form
\begin{equation}
\mathbf{Y}_c = \sum_{c'=1}^{C_{\rm in}} w_{cc'} \mathbf{X}_{c'} ,
\end{equation}
where \(w_{cc'}\in\mathbb R\). Since the same scalar is applied to all three coordinate components, the operation commutes with rotations.

Temporal convolution is applied analogously with scalar kernels over time and channels:
\begin{equation}
\mathbf{Y}_{t,c} = \sum_{\tau,c'} k_{\tau c c'} \mathbf{X}_{t-\tau,c'} .
\end{equation}
Nonlinearities gate equivariant vectors using invariant scalars such as norms, inner products, and Lie-bracket/cross-product interactions. These operations preserve the \(\mathrm{SO}(3)\) action and are used inside residual temporal blocks.

\subsection{Structural Parameter Reduction}
\label{app:param_efficiency}

The parameter reduction of GINIO is structural. For a temporal layer with kernel size \(K\), \(C_{\rm in}\) input vector channels, and \(C_{\rm out}\) output vector channels, an unrestricted scalar-coordinate convolution over flattened coordinates has
\begin{equation}
K \times 3C_{\rm in} \times 3C_{\rm out}
=
9K C_{\rm in}C_{\rm out}
\end{equation}
weights. An equivariant vector-channel convolution instead uses
\begin{equation}
K \times C_{\rm in}\times C_{\rm out}
\end{equation}
scalar weights, sharing the learned interaction across the geometric \(\mathbb R^3\) components. Each channel-mixing block is thus \(9\times\) smaller by construction, which is why the ResNet-style GINIO-RD temporal backbones use 0.57M parameters versus 5.4M for non-equivariant ResNet-style baselines.

\subsection{Numerical Equivariance Verification}
\label{app:equivariance_verification}

We verify equivariance by sampling random rotations \(\mathbf{R}\in\mathrm{SO}(3)\), rotating all vector-valued inputs, rerunning the model, and comparing the result to the rotated original prediction. For a vector output \(\boldsymbol{\mu}\), the error is
\begin{equation}
\|\boldsymbol{\mu}(\rho(\mathbf{R})\mathbf{x})-\mathbf{R}\boldsymbol{\mu}(\mathbf{x})\|_2,
\end{equation}
and for covariance,
\begin{equation}
\|\boldsymbol{\Sigma}(\rho(\mathbf{R})\mathbf{x})-\mathbf{R}\boldsymbol{\Sigma}(\mathbf{x})\mathbf{R}^\top\|_F.
\end{equation}

The AirIO-style GINIO-A recurrent components---vector-channel linear layers, nonlinearities, layer normalization, and recurrent cells---satisfy equivariance errors between \(10^{-15}\) and \(10^{-13}\) in float64. The full GINIO-A model accumulates to a mean-equivariance error below \(5\times10^{-9}\). Its diagonal covariance head is not claimed to be tensor-equivariant, consistent with the covariance analysis in the main text. The GINIO-E full-\(\mathrm{SO}(3)\) canonicalized model satisfies both mean and covariance equivariance to within \(10^{-15}\).

\subsection{GINIO Instantiations}
\label{app:ginio_instantiations}

Table~\ref{tab:ginio_instantiations} summarizes the suffix convention used throughout the paper. All variants share the same $\mathrm{SO}(3)$-equivariant motion interface; variants equipped with tensor-equivariant uncertainty heads additionally satisfy the covariance congruence law in Eq.~(3). Diagonal covariance heads are retained as explicit uncertainty ablations. The suffixes indicate the experimental backend, frame convention, or uncertainty head.

\begin{table}[t]
\centering
\small
\caption{GINIO suffix convention.}
\label{tab:ginio_instantiations}
\setlength{\tabcolsep}{5pt}
\begin{tabular}{lll}
\toprule
Suffix & Experiment & Instantiation \\
\midrule
-W  & TLIO/RIDI  & world-frame equivariant temporal model \\
-S  & TLIO/RIDI  & sensor-frame equivariant temporal model \\
-A  & NanoBench & AirIO-style equivariant recurrent model \\
-E  & NanoBench & full-\(\mathrm{SO}(3)\) canonicalized model \\
-RD & NanoBench & equivariant temporal model with diagonal uncertainty \\
-RS & NanoBench & equivariant temporal model with spectral uncertainty \\
-R  & Stress tests / Fetch & ResNet-style equivariant temporal model \\
\bottomrule
\end{tabular}
\end{table}

\subsection{Equivariant Reinstantiation of Heterogeneous Backbones}
\label{app:equiv_reinstantiation}

A central design claim of GINIO is that the same SO(3) interface can be realized by fully
equivariant reconstructions of architecturally \emph{heterogeneous} inertial backbones,
rather than by a single bespoke network. We demonstrate this by reinstantiating three
families that differ in their temporal mechanism and in their original symmetry assumptions,
none of which is fully \(\mathrm{SO}(3)\)-equivariant in its published form:
\begin{itemize}
\item a \textbf{ResNet-style temporal} backbone (GINIO-RD/RS), whose original form treats
the three coordinate components as ordinary scalar channels;
\item an \textbf{AirIO-style recurrent} backbone (GINIO-A), whose original form uses a
bidirectional GRU with direct, non-equivariant gate mixing; and
\item an \textbf{EqNIO-style canonicalized} backbone (GINIO-E), whose original form enforces
only the gravity-stabilizer subgroup \(O_g(3)\cong O(2)\).
\end{itemize}
In every case we replace the learned temporal, recurrent, canonicalization, and motion-prediction components with operations built from
the same ReLN equivariant primitives, so that the vector-valued learned motion computation, not merely a
sub-block, satisfies the full $\mathrm{SO}(3)$ transformation law. Where tensor-equivariant uncertainty is required, we use the spectral covariance construction to satisfy the congruence law in Eq.~(3); diagonal covariance heads are retained only as explicit uncertainty ablations. Two of these
reconstructions are, to our knowledge, the first fully SO(3)-equivariant formulations of
their respective backbones: an equivariant gated recurrent cell for inertial prediction
(\S\ref{app:ginio_a_details}) and a full-SO(3) canonicalized predictor that lifts the
EqNIO-style \(O(2)\) design to the entire rotation group (\S\ref{app:ginio_e_details}). The
per-family equivariance errors in \S\ref{app:equivariance_verification} confirm that each
instantiation is equivariant to numerical precision, not only approximately.

\subsection{GINIO-R: ResNet-Style Equivariant Temporal Instantiation}
\label{app:ginio_r_details}

GINIO-RD and GINIO-RS instantiate the GINIO interface in a ResNet-style temporal inertial network. The non-equivariant ResNet treats the three coordinate components as ordinary scalar channels. GINIO-R instead represents each feature as vector channels in \(\mathbb{R}^3\), and replaces scalar-coordinate temporal blocks with \(\mathrm{SO}(3)\)-equivariant ReLN-style temporal blocks~\citep{kim2025equivariant}. Channel mixing and temporal convolution use scalar weights over vector channels, while nonlinearities are built from rotation-invariant gates and equivariant vector interactions such as cross products. Every intermediate vector feature therefore transforms by the standard \(\mathrm{SO}(3)\) action.

GINIO-RD and GINIO-RS share this equivariant temporal backbone. GINIO-RD uses a diagonal covariance head and is the accuracy-oriented variant for fixed-frame trajectory prediction. GINIO-RS uses the spectral covariance head and is the geometry-consistent uncertainty variant when covariance itself is consumed as a rotating tensor.

\subsection{GINIO-A: AirIO-Style Recurrent Instantiation}
\label{app:ginio_a_details}

GINIO-A instantiates the GINIO interface in an AirIO-style aerial prediction pipeline. The
original AirIO architecture encodes a body-frame IMU window with a 1-D convolutional
network and feeds the features to a bidirectional GRU that regresses body-frame velocity.
We make each learned component \(\mathrm{SO}(3)\)-equivariant using the equivariant linear
and nonlinearity primitives of Reductive Lie Neurons (ReLN)~\citep{kim2025equivariant} as
building blocks for the convolutional encoder, the recurrent cell, and the output decoder.
The accelerometer and gyroscope enter as two vector channels in \(\mathbb{R}^3\), and the
encoder applies scalar temporal kernels shared across the geometric components, so the same
filter acts on all spatial axes.

The recurrent cell is where the AirIO GRU is made equivariant, so we describe it in detail.
Its building blocks are the ReLN equivariant linear
\(\mathrm{Lin}_{\mathbf{W}}(\mathbf{U})=\mathbf{W}\mathbf{U}\), which mixes channels but
never acts on the \(\mathbb{R}^3\) axis, and the ReLN equivariant nonlinearity \(\phi\),
together with an equivariant LayerNorm. The hidden state is a stack of vector channels
\(\mathbf{H}_t\in\mathbb{R}^{C_h\times 3}\). A standard GRU forms its gates by mixing input
and hidden features directly, which is not rotation-invariant; we instead derive the gates
from invariant scalars. Projecting the vectors with an equivariant map \(\mathbf{W}_n\) and
taking per-channel norms gives \(s(\mathbf{U})=(\|(\mathbf{W}_n\mathbf{U})_i\|)_i\),
invariant because \(\|\mathbf{R}\mathbf{v}\|=\|\mathbf{v}\|\). The reset and update gates
\(\mathbf{r}_t,\mathbf{u}_t=\sigma(g_{r,u}([s(\mathbf{X}_t),s(\mathbf{H}_{t-1})]))\in[0,1]^{C_h}\)
are then rotation-invariant, and the cell updates as
\begin{equation}
\tilde{\mathbf{H}}_t
=
\mathrm{LN}\Big(\phi\big(
\mathrm{Lin}_{\mathbf{W}_x}(\mathbf{X}_t)+\mathrm{Lin}_{\mathbf{W}_h}(\mathbf{r}_t\odot\mathbf{H}_{t-1})
\big)\Big),
\qquad
\mathbf{H}_t=(1-\mathbf{u}_t)\odot\mathbf{H}_{t-1}+\mathbf{u}_t\odot\tilde{\mathbf{H}}_t,
\end{equation}
where \(\odot\) broadcasts each scalar gate onto the three components of its channel. Every
operation either mixes channels, acts on invariant norms, or multiplies vector channels by
invariant scalars, so rotating all inputs by \(\mathbf{R}\) rotates every hidden state by
the same \(\mathbf{R}\); the equivariant decoder then maps to a single \(\mathbb{R}^3\)
velocity, giving \(\hat{\mathbf{m}}\mapsto\mathbf{R}\hat{\mathbf{m}}\). We stack the cells
bidirectionally, concatenating forward and backward outputs along the vector-channel axis. To our knowledge, this is the first fully
SO(3)-equivariant gated recurrent cell applied to neural inertial odometry.

We evaluate two input settings. In the no-attitude setting, both AirIO and GINIO-A receive
only body-frame accelerometer and gyroscope channels; this input-matched comparison
isolates the equivariant architecture prior. In the with-attitude setting, we do not use
\(\operatorname{Log}(\mathbf{R}_{WS})^\vee\) directly, since a raw attitude log-map does not in general transform as a vector under a sensor-frame right action. Instead, let \(\mathbf{e}_i\) be the standard basis and define \(\mathbf{D}_i=\operatorname{Exp}([\mathbf{e}_i]_\times)\). We encode attitude using
\begin{equation}
\mathbf{f}_i(\mathbf{R}_{WS})
=\operatorname{Log}\!\left(\mathbf{R}_{WS}^{\mathsf T}\mathbf{D}_i\mathbf{R}_{WS}\right)^\vee
=\mathbf{R}_{WS}^{\mathsf T}\mathbf{e}_i, \qquad i=1,2,3.
\end{equation}
Under a sensor-frame rotation \(\mathbf{R}_{WS}\mapsto\mathbf{R}_{WS}\mathbf{Q}^{\mathsf T}\), this encoding obeys \(\mathbf{f}_i\mapsto\mathbf{Q}\mathbf{f}_i\). The three \(\mathbf{f}_i\) are therefore supplied as additional equivariant vector channels, making the attitude-aided input compatible with the same \(\mathrm{SO}(3)\) action as the IMU vectors.

\subsection{GINIO-E: EqNIO-Style Full-\texorpdfstring{\(\mathrm{SO}(3)\)}{SO(3)} Canonicalized Instantiation}
\label{app:ginio_e_details}

GINIO-E is our full-\(\mathrm{SO}(3)\) counterpart to the EqNIO-style canonicalized architecture family. EqNIO enforces \(O_g(3)\cong O(2)\) subequivariance, which is appropriate when the input is gravity-stabilized and the remaining symmetry is rotation about the gravity axis. Arbitrary IMU remounting, however, acts by the full \(\mathrm{SO}(3)\) group on calibrated sensor-frame vectors.

GINIO-E keeps the canonicalization-and-prediction structure but replaces the gravity-stabilizer frame with a full \(\mathrm{SO}(3)\)-equivariant frame. The frame predictor outputs two equivariant \(3\)-D vectors. We normalize the first, remove its projection from the second, normalize the residual, and form the third basis vector with a cross product. This produces a right-handed frame \(\mathbf{R}_F\in\mathrm{SO}(3)\) whenever the two predicted vectors are non-collinear; near-degenerate cases are stabilized by norm clamping during normalization. This frame construction is used for canonicalization and is distinct from the spectral covariance head, whose orthonormal frame uses the polar/SVD parameterization of Appendix~\ref{app:cov_details}.

The input window is canonicalized by applying \(\mathbf{R}_F^\top\) to the vector axis. Since the frame transforms as \(\mathbf{R}_F\mapsto \mathbf{Q} \mathbf{R}_F\) under an external rotation \(\mathbf{Q}\), the canonicalized input is invariant to \(\mathbf{Q}\), and a standard temporal backbone can operate in the canonical frame. The predicted displacement and covariance are then mapped back to the original frame:
\begin{equation}
    \hat{\mathbf{m}} = \mathbf{R}_F \hat{\mathbf{m}}_{\rm can},
    \qquad
    \hat{\boldsymbol{\Sigma}} =
    \mathbf{R}_F \operatorname{diag}(\boldsymbol{\sigma}^2_{\rm can}) \mathbf{R}_F^\top ,
\end{equation}
which yields
\begin{equation}
\hat{\mathbf{m}}(\rho(\mathbf{Q})\bar{\mathcal Z})=
\mathbf{Q}\hat{\mathbf{m}}(\bar{\mathcal Z}), \qquad \hat{\boldsymbol{\Sigma}}(\rho(\mathbf{Q})\bar{\mathcal Z}) =
\mathbf{Q}\hat{\boldsymbol{\Sigma}}(\bar{\mathcal Z})\mathbf{Q}^\top.
\end{equation}
We include GINIO-E to show that the GINIO interface is not tied to a single backbone family but can also be realized in a canonicalized architecture; we use the suffix E to distinguish it from the original EqNIO O(2) model. GINIO-E thus lifts the canonicalization-based EqNIO design from its gravity-stabilizer \(O(2)\) symmetry to the full \(\mathrm{SO}(3)\) group: the learned canonical frame is itself \(\mathrm{SO}(3)\)-equivariant, so the predictor is no longer tied to a privileged gravity axis.

\subsection{AirIO-Style Capacity Ablation on NanoBench}
\label{app:ginio_a_capacity}

The channel-reduced GINIO-A controls for the possibility that the AirIO-style equivariant result is merely due to larger capacity. Table~\ref{tab:app_ginio_a_capacity} shows that a \(6.1\times\) smaller recurrent equivariant model slightly improves ATE over the original GINIO-A, and that the reduced model is smaller than both AirIO and the original equivariant recurrent variant.

\begin{table}[htbp]
\centering
\small
\caption{AirIO-style GINIO-A capacity ablation on NanoBench. The reduced model has \(6.1\times\) fewer parameters than the original equivariant recurrent model and slightly improves ATE.}
\label{tab:app_ginio_a_capacity}
\setlength{\tabcolsep}{5pt}
\begin{tabular}{lccc}
\toprule
Variant & Input & Params & ATE [m] \(\downarrow\) \\
\midrule
GINIO-A original & body & 0.273M & 1.502 \\
GINIO-A reduced & body & \textbf{0.045M} & \textbf{1.430} \\
GINIO-A original & body+att. & 0.273M & 1.062 \\
GINIO-A reduced & body+att. & \textbf{0.045M} & \textbf{1.045} \\
\bottomrule
\end{tabular}
\end{table}

This supports the view that equivariance acts as a structural regularizer in the aerial setting, rather than as added representational capacity.

\section{Proof of Last-Frame Alignment Equivalence}
\label{app:lfa_proof}

\begin{figure}[htbp]
\centering
\includegraphics[width=0.8\linewidth]{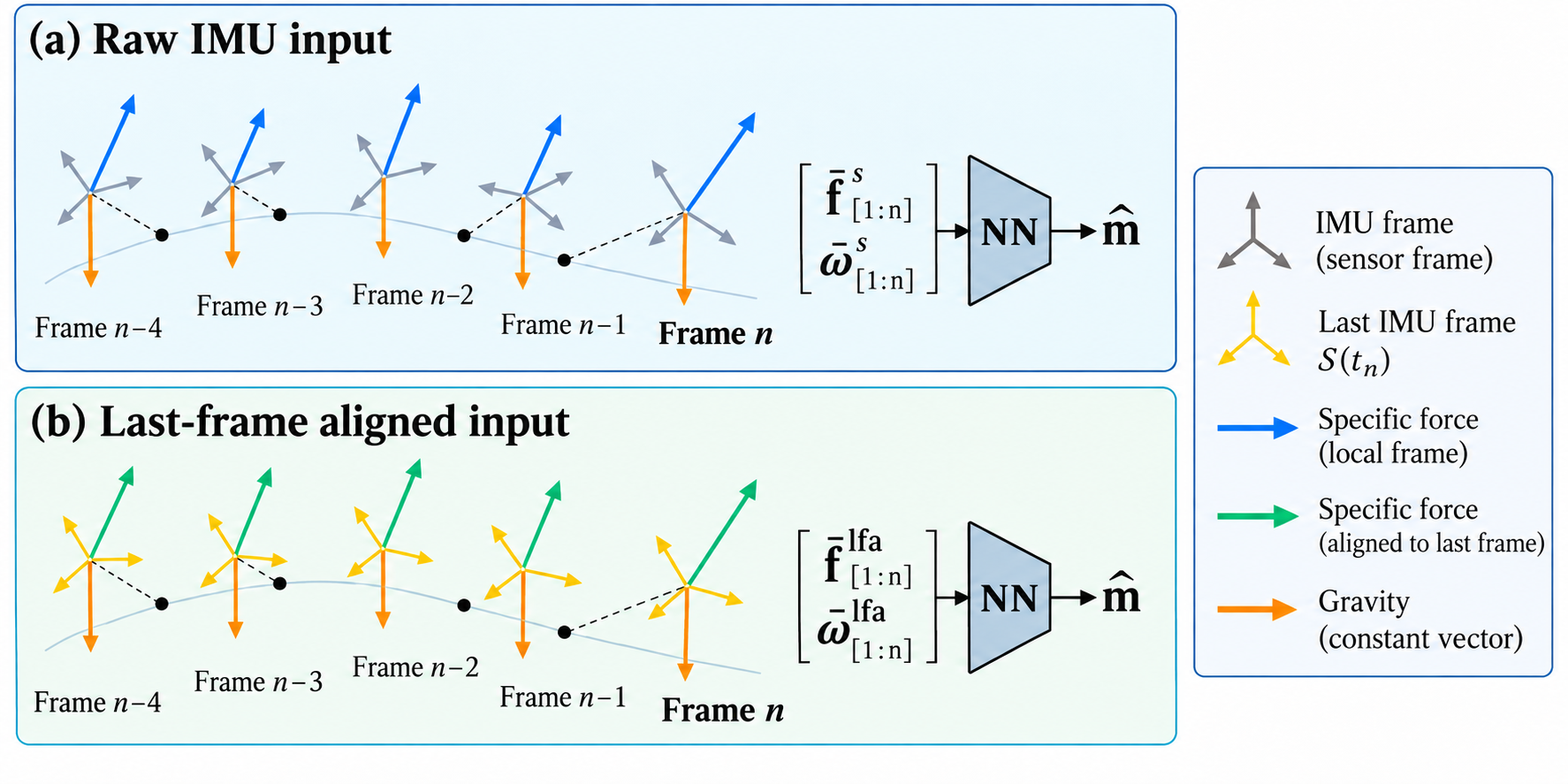}
\caption{Last-Frame Alignment (LFA). (a) In the raw IMU window, each sample is expressed in its instantaneous sensor frame \(\mathcal{S}(t_k)\), whose orientation varies over time. (b) LFA rotates all samples into the last sensor frame \(\mathcal{S}(t_n)\), yielding a common coordinate system over the input window. Both representations are processed by the same network with the same target; LFA modifies only the input coordinates.}
\label{fig:app_lfa}
\end{figure}

Figure~\ref{fig:app_lfa} illustrates the intuition behind LFA before we state the formal
equivalence. In raw sensor-frame windows (Fig.~\ref{fig:app_lfa}a), the gravity component rotates within the
window as the sensor attitude changes, producing a time-varying offset in the input. LFA (Fig.~\ref{fig:app_lfa}b) rotates every sample into the sensor frame at the final timestep, which renders the
gravity direction approximately constant across the window and reduces within-window
rotational nonstationarity, while keeping the representation sensor-centric.

Let \({}^{w}_{s}\mathbf{R}_k\) be the rotation from the sensor frame \(\mathcal S(t_k)\) to the world frame. The world-frame sequence is
\begin{equation}
\bar{\mathcal Z}^{w}
=
\left\{
\rho({}^{w}_{s}\mathbf{R}_k)\bar{\mathbf{z}}_k^s
\right\}_{k=1}^{n}.
\end{equation}
LFA rotates every sample into the final sensor frame \(\mathcal S(t_n)\):
\begin{equation}
\mathbf{R}_{n,k}
=
({}^{w}_{s}\mathbf{R}_n)^\top {}^{w}_{s}\mathbf{R}_k,
\qquad
\bar{\mathbf{z}}_k^{\rm lfa}
=
\rho(\mathbf{R}_{n,k})\bar{\mathbf{z}}_k^s .
\end{equation}
Thus,
\begin{equation}
\bar{\mathcal Z}^{\rm lfa}
=
\left\{
\rho(({}^{w}_{s}\mathbf{R}_n)^\top{}^{w}_{s}\mathbf{R}_k)\bar{\mathbf{z}}_k^s
\right\}_{k=1}^{n}
=
\rho(({}^{w}_{s}\mathbf{R}_n)^\top)\bar{\mathcal Z}^{w}.
\end{equation}
If \(\Phi\) is \(\mathrm{SO}(3)\)-equivariant, then
\begin{equation}
\Phi(\bar{\mathcal Z}^{\rm lfa})
=
\Phi(\rho(({}^{w}_{s}\mathbf{R}_n)^\top)\bar{\mathcal Z}^{w})
=
({}^{w}_{s}\mathbf{R}_n)^\top \Phi(\bar{\mathcal Z}^{w}).
\end{equation}
Rearranging gives
\begin{equation}
\Phi(\bar{\mathcal Z}^{w})
=
{}^{w}_{s}\mathbf{R}_n \Phi(\bar{\mathcal Z}^{\rm lfa}).
\end{equation}
For an \(\mathrm{SO}(3)\)-equivariant predictor, LFA and world-frame training therefore differ only by the final rotation: LFA is a sensor-frame representation choice, and the physical target is unchanged.

For a covariance-valued head satisfying the congruence equivariance law, the same change-of-frame argument gives
\begin{equation}
\hat{\boldsymbol{\Sigma}}(\bar{\mathcal Z}^{w})
=
{}^{w}_{s}\mathbf{R}_n
\hat{\boldsymbol{\Sigma}}(\bar{\mathcal Z}^{\rm lfa})
({}^{w}_{s}\mathbf{R}_n)^\top .
\end{equation}
Last-Frame Alignment thus preserves both the vector output law and the covariance congruence law.

\section{EqNIO Compatibility: \texorpdfstring{\(O_g(3)\)}{Og(3)} vs. Full \texorpdfstring{\(\mathrm{SO}(3)\)}{SO(3)}}
\label{app:compatibility}

EqNIO is the closest architecture-level rotation-aware NIO baseline. Its symmetry is the stabilizer of the gravity direction,
\begin{equation}
O_g(3)=\{\mathbf{R}\in O(3):\mathbf{R}\mathbf{g}=\mathbf{g}\}\cong O(2),
\end{equation}
which corresponds to yaw rotations and reflections around the gravity axis in a gravity-stabilized frame. This is a useful prior when the input is accurately gravity-aligned. An arbitrary three-dimensional change of sensor-frame convention is not restricted to the gravity stabilizer. Its coordinate-frame action is the full \(\mathrm{SO}(3)\) group, under which the learned vector/covariance pair must transform as
\begin{equation}
(\hat{\mathbf{m}},\hat{\boldsymbol{\Sigma}})
\mapsto
(\mathbf{R}\hat{\mathbf{m}},\mathbf{R}\hat{\boldsymbol{\Sigma}}\mathbf{R}^\top).
\end{equation}
GINIO imposes this full \(\mathrm{SO}(3)\)-equivariance on the learned computation. This is why EqNIO can be competitive in nominal gravity-stabilized settings but degrades more than a full-\(\mathrm{SO}(3)\) predictor under out-of-plane remounting and under frame-estimation mismatch outside its assumed symmetry group.

\section{Additional Human-Motion Metrics and Controlled Ablations}
\label{app:full_metrics}

\subsection{Trajectory Metric Definitions}
\label{app:metrics_arxiv}
For a trajectory with positions \(\mathbf p_i\) and estimates \(\hat{\mathbf p}_i\), ATE is the RMSE of aligned positions over the full trajectory. Following the TLIO evaluation protocol, relative translation error is computed over fixed \(\Delta t=1\,\mathrm{s}\) windows. Let \(R_{\gamma,j}\) and \(\hat R_{\gamma,j}\) denote the ground-truth and estimated yaw rotations at the start of window \(j\). Then
\begin{equation}
\mathrm{RTE}(\Delta t)
=\left[
\frac{1}{M}\sum_{j=1}^{M}
\left\|
(\mathbf p_{j+\Delta t}-\mathbf p_j)
-
R_{\gamma,j}\hat R_{\gamma,j}^{\mathsf T}
(\hat{\mathbf p}_{j+\Delta t}-\hat{\mathbf p}_j)
\right\|_2^2
\right]^{1/2},
\end{equation}
where \(M\) is the number of valid windows. The yaw alignment removes accumulated heading error at the beginning of each window so that RTE emphasizes local translational consistency. Drift ratio is
\begin{equation}
\mathrm{DR}=100\,\frac{\|\mathbf p_N-\hat{\mathbf p}_N\|_2}
{\sum_{i=1}^{N-1}\|\mathbf p_{i+1}-\mathbf p_i\|_2}.
\end{equation}

\subsection{Full TLIO/RIDI Metrics}

Table~\ref{tab:app_human_full_metrics} reports the full trajectory metrics available for TLIO and RIDI. ATE is the mean trajectory error used in the main text; RTE and drift ratio provide local consistency and normalized long-horizon drift diagnostics.

\begin{table*}[t]
\centering
\small
\caption{Full TLIO/RIDI metrics. RTE and drift ratio are reported from the matched native evaluation protocols; entries marked ``--'' were not available for the corresponding updated run. Here, aug.\ denotes full \(\mathrm{SO}(3)\) train-time augmentation.}
\label{tab:app_human_full_metrics}
\renewcommand{\arraystretch}{1.08}
\begin{tabular*}{\textwidth}{@{\extracolsep{\fill}}l l rrr rrr@{}}
\toprule
\multirow{2}{*}{Dataset} & \multirow{2}{*}{Method}
& \multicolumn{3}{c}{ID/ID}
& \multicolumn{3}{c@{}}{ID/\(\mathrm{SO}(3)\)} \\
\cmidrule(lr){3-5}\cmidrule(l){6-8}
& & ATE & RTE & DR & ATE & RTE & DR \\
\midrule
\multirow{6}{*}{TLIO}
& ResNet+SCEKF & 1.725 & 0.126 & 1.76 & 35.699 & 0.565 & 55.45 \\
& ResNet+SCEKF+aug. & 1.844 & 0.133 & 1.87 & 8.687 & 0.565 & 46.26 \\
& EqNIO & 1.554 & 0.120 & 1.32 & 76.389 & 0.723 & 41.46 \\
& GINIO-W & 1.615 & 0.123 & 1.37 & 2.018 & 0.145 & 2.30 \\
& ResNet-S+InEKF & 2.814 & 0.177 & 3.83 & 36.593 & 0.579 & 57.20 \\
& GINIO-S & 2.704 & 0.173 & 3.54 & 2.759 & 0.207 & 3.70 \\
\midrule
\multirow{6}{*}{RIDI}
& ResNet+SCEKF & 1.207 & 0.107 & 1.45 & 2.990 & 0.275 & 2.14 \\
& ResNet+SCEKF+aug. & 1.401 & 0.147 & 1.63 & 3.011 & 0.281 & 2.41 \\
& EqNIO & 2.408 & 0.188 & 2.66 & 3.460 & 0.219 & 2.89 \\
& GINIO-W & 0.952 & 0.109 & 1.70 & 2.450 & 0.253 & 1.99 \\
& ResNet-S+InEKF & 1.445 & 0.151 & 1.87 & 4.752 & 0.286 & 3.46 \\
& GINIO-S & 1.319 & 0.150 & 1.83 & 1.457 & 0.152 & 2.01 \\
\bottomrule
\end{tabular*}
\end{table*}

\subsection{Controlled TLIO Network Ablation}
\label{app:ablation_full}

Table~\ref{tab:app_tlio_controlled_ablation} isolates architectural equivariance, augmentation, and LFA in a controlled TLIO setting. \(\mathrm{SO}(3)\) augmentation improves the non-equivariant baseline from 108.2 to 3.2 under ID/\(\mathrm{SO}(3)\) but does not impose the continuous transformation law, so the equivariant model remains more stable. LFA restores the sensor-frame model to world-frame accuracy when the predictor is equivariant, matching the LFA equivalence in Appendix~\ref{app:lfa_proof}.

\begin{table}[htbp]
\centering
\small
\caption{Controlled TLIO ablation. ATE [m] under ID/ID and ID/\(\mathrm{SO}(3)\).}
\label{tab:app_tlio_controlled_ablation}
\setlength{\tabcolsep}{5pt}
\begin{tabular}{lcc}
\toprule
Model & ID/ID & ID/\(\mathrm{SO}(3)\) \\
\midrule
ResNet & 1.817 & 108.235 \\
ResNet+\(\mathrm{SO}(3)\) aug. & 1.966 & 3.189 \\
GINIO-Sensor & 2.293 & 2.293 \\
GINIO-World & \textbf{1.828} & \textbf{1.828} \\
GINIO-Sensor+LFA & \textbf{1.828} & \textbf{1.828} \\
\bottomrule
\end{tabular}
\end{table}

\subsection{Matched Backend and Filter Ablations}

Table~\ref{tab:app_backend_ablation} shows that rotational robustness is not a filter artifact. With the same InEKF backend, the non-equivariant ResNet-S diverges under ID/\(\mathrm{SO}(3)\), whereas GINIO-S remains stable.

\begin{table}[htbp]
\centering
\small
\caption{Sensor-frame system ablations on TLIO. ATE [m].}
\label{tab:app_backend_ablation}
\setlength{\tabcolsep}{5pt}
\begin{tabular}{llccc}
\toprule
Frontend & Filter & Gravity comp. & ID/ID & ID/\(\mathrm{SO}(3)\) \\
\midrule
ResNet & InEKF & \(\times\) & 2.814 & 36.593 \\
GINIO & InEKF & \(\times\) & \textbf{2.704} & \textbf{2.759} \\
\midrule
GINIO & SCEKF & \(\checkmark\) & 4.191 & 6.455 \\
GINIO & InEKF & \(\checkmark\) & 2.883 & 3.424 \\
\midrule
GINIO & InEKF & \(\checkmark\) & 2.883 & 3.424 \\
GINIO & InEKF & \(\times\) & \textbf{2.704} & \textbf{2.759} \\
\bottomrule
\end{tabular}
\end{table}

\section{NanoBench Details and Additional Results}
\label{app:nanobench}

\subsection{Protocol and Metrics}

NanoBench is evaluated as a neural-network-only aerial prediction benchmark over 17
held-out sequences, with ATE computed after the benchmark alignment protocol. The network
inputs are the accelerometer and gyroscope vector channels: the no-attitude setting uses
body-frame IMU only, and the with-attitude setting additionally provides the equivariant
attitude vector-channel encoding defined in Appendix~\ref{app:ginio_a_details}. For the ID/\(\mathrm{SO}(3)\) test we rotate all vector-valued
inputs by a sampled \(\mathbf{R}\in\mathrm{SO}(3)\) and unrotate the predictions before
computing metrics. Beyond ATE, we report a velocity-RMSE diagnostic used for the
rotation-stress evaluation.

\subsection{NanoBench ID/\texorpdfstring{\(\mathrm{SO}(3)\)}{SO(3)} Velocity Stress}

\begin{figure}[htbp]
\centering
\includegraphics[width=0.72\textwidth]{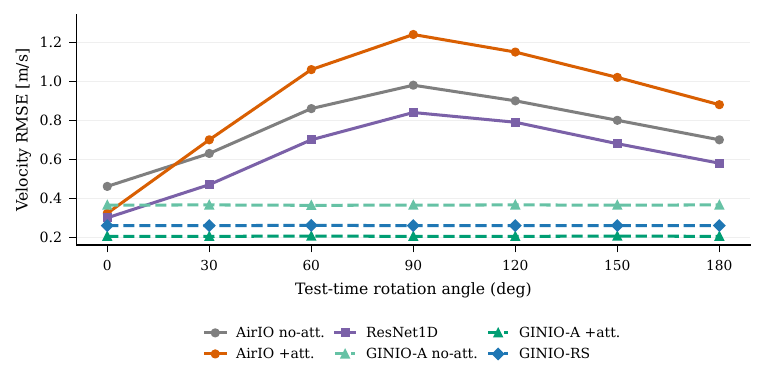}
\caption{NanoBench ID/\(\mathrm{SO}(3)\) velocity-RMSE stress test. Non-equivariant models vary with the input frame, while GINIO variants remain stable across rotations.}
\label{fig:app_nanobench_velocity_stress}
\end{figure}

Figure~\ref{fig:app_nanobench_velocity_stress} reports velocity-RMSE under test-time frame rotations. Non-equivariant AirIO and ResNet1D vary with the input frame, whereas GINIO-A and GINIO-RS remain stable across rotations.

\section{Spectral Covariance Details}
\label{app:cov_details}

\subsection{Equivariant SPD Parameterization}
GINIO predicts a covariance by combining an equivariant eigenbasis and invariant
eigenvalues. Given three equivariant vectors, we form
\begin{equation}
\mathbf{A}=[\mathbf{a}_1,\mathbf{a}_2,\mathbf{a}_3]\in\mathbb R^{3\times 3}.
\end{equation}
We compute the SVD \(\mathbf{A}=\mathbf{U}\mathbf{S}\mathbf{W}^\top\) and use the polar projection
\begin{equation}
\mathbf{V} = \mathbf{U}\,\mathrm{diag}(1,1,\det(\mathbf{U}\mathbf{W}^\top))\,\mathbf{W}^\top \in \mathrm{SO}(3).
\end{equation}
The network separately predicts invariant log-standard deviations \(\lambda_i\), and the
covariance is
\begin{equation}
\hat{\boldsymbol{\Sigma}}
=
\mathbf{V}\,\mathrm{diag}(\exp(2\lambda_1),\exp(2\lambda_2),\exp(2\lambda_3))\,\mathbf{V}^\top .
\end{equation}
Under an input rotation \(\mathbf{R}\), the equivariant vectors satisfy \(\mathbf{A}\mapsto \mathbf{R}\mathbf{A}\), so the
orthonormalized frame satisfies \(\mathbf{V}\mapsto \mathbf{R}\mathbf{V}\) whenever \(\mathbf{A}\) is full rank, while the
eigenvalues remain invariant. Therefore
\begin{equation}
\hat{\boldsymbol{\Sigma}}(\rho(\mathbf{R})\bar{\mathcal Z})
=
\mathbf{R}\hat{\boldsymbol{\Sigma}}(\bar{\mathcal Z})\mathbf{R}^\top .
\end{equation}
Near-degenerate cases, where the predicted vectors are nearly collinear, are stabilized
by numerical regularization during orthonormalization; the resulting equivariance is
confirmed by the covariance-equivariance error reported in
Table~\ref{tab:app_cov_ablation_full}.

\subsection{Training Losses for Motion and Uncertainty}
For a target residual \(\mathbf{e} = \mathbf{y}-\hat{\mathbf{m}}\), the Gaussian negative
log-likelihood is
\begin{equation}
\mathcal L_{\rm NLL}
=
\frac{1}{2}\mathbf{e}^\top \hat{\boldsymbol{\Sigma}}^{-1}\mathbf{e}
+
\frac{1}{2}\log\det \hat{\boldsymbol{\Sigma}},
\end{equation}
and the MSE baseline uses
\begin{equation}
\mathcal L_{\rm MSE}=\|\mathbf{y}-\hat{\mathbf{m}}\|_2^2.
\end{equation}
For spectral covariance, the NLL is computed with the SPD covariance
\(\hat{\boldsymbol{\Sigma}}=\mathbf{V}\,\mathrm{diag}(\sigma_1^2,\sigma_2^2,\sigma_3^2)\,\mathbf{V}^\top\).

\subsection{Covariance Metrics}

We evaluate:
\begin{itemize}
    \item velocity RMSE for prediction accuracy;
    \item NLL and its Mahalanobis/log-determinant decomposition;
    \item NEES and empirical coverage for uncertainty calibration;
    \item covariance equivariance error
    \begin{equation}
    \frac{\|\hat{\boldsymbol{\Sigma}}(\rho(\mathbf{R})\bar{\mathcal Z})-\mathbf{R}\hat{\boldsymbol{\Sigma}}(\bar{\mathcal Z})\mathbf{R}^\top\|_F}
    {\|\mathbf{R}\hat{\boldsymbol{\Sigma}}(\bar{\mathcal Z})\mathbf{R}^\top\|_F};
    \end{equation}
    \item motion equivariance error for the learned mean.
\end{itemize}

\subsection{Full Covariance Ablation Table}
Table~\ref{tab:app_cov_ablation_full} reports the raw covariance ablation. The main
text focuses on NLL-trained diagonal vs.\ spectral heads; here we include the
MSE-trained variants as well.
\begin{table*}[htbp]
\centering
\small
\caption{NanoBench covariance ablation. Lower is better for RMSE, NLL, and equivariance errors. Dashes indicate
uncertainty-calibration metrics that are not meaningful for MSE-trained heads because
MSE does not optimize a probabilistic covariance.}
\label{tab:app_cov_ablation_full}
\begin{tabular*}{\textwidth}{@{\extracolsep{\fill}}llcccccc@{}}
\toprule
Head & Loss & Vel. RMSE & NLL & NEES & cov95 & \(\boldsymbol{\Sigma}\)-eq. err. & \(\boldsymbol{\mu}\)-eq. err. \\
\midrule
Diag     & MSE & 0.284 & --     & --    & --    & --                  & 0.453 \\
Diag     & NLL & 0.303 & -1.741 & 2.959 & 0.915 & \(5.52{\times}10^{-1}\) & 0.317 \\
Spectral & MSE & 0.234 & --     & --    & --    & --                  & \(7.24{\times}10^{-4}\) \\
Spectral & NLL & 0.260 & -1.875 & 2.974 & 0.837 & \(4.34{\times}10^{-4}\) & \(4.30{\times}10^{-4}\) \\
\bottomrule
\end{tabular*}
\end{table*}

\subsection{Uncertainty Calibration}
\label{app:cov-cal}
\begin{figure}[t]
  \centering
  \includegraphics[width=0.9\linewidth]{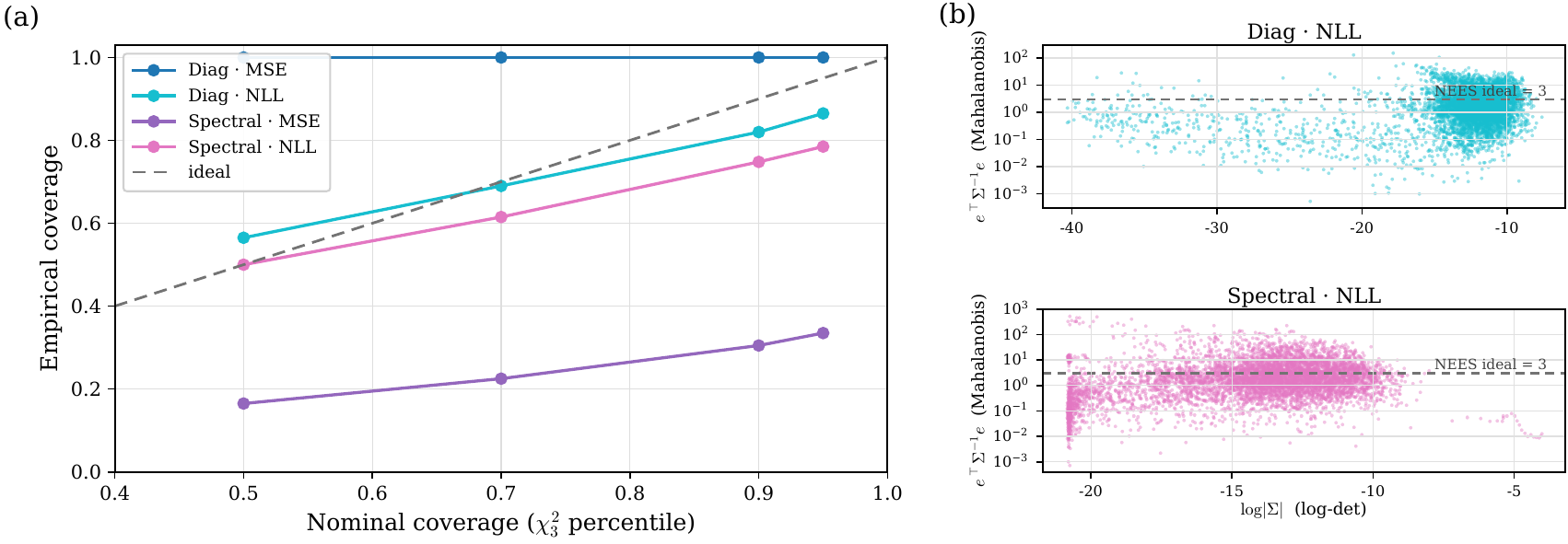}
  \caption{(a) Empirical NEES coverage on the NanoBench evaluation. MSE-trained heads are poorly calibrated (the diagonal head saturates near
  \(1\); the spectral head is over-confident); the two NLL-trained heads track the ideal
  more closely, with the \(\mathrm{Spectral}\cdot\mathrm{NLL}\) head closest to ideal in
  the tail. (b) Per-window NLL decomposition for the NLL-trained heads. The dashed line
  marks the ideal NEES \(=3\). The spectral head spans a wider range of
  log-determinants, indicating input-dependent (heteroscedastic) uncertainty, whereas the
  diagonal head is concentrated at near-constant scale.}
  \label{fig:cov-cal}
\end{figure}

We report the raw NEES coverage of each covariance head, with no post-hoc scaling,
where \(\mathrm{NEES}=\mathbf{e}^\top\hat{\boldsymbol{\Sigma}}^{-1}\mathbf{e}\) has ideal
value equal to the state dimension (Fig.~\ref{fig:cov-cal}a). The MSE-trained heads are
poorly calibrated: the diagonal head is under-confident (its coverage saturates near
\(1\)) and the spectral head is over-confident, consistent with MSE not learning
input-dependent uncertainty. The two NLL-trained heads are better calibrated and
comparable on aggregate \(\mathrm{cov}_{95}\) (\(0.915\) for diagonal, \(0.837\) for
spectral; Table~\ref{tab:app_cov_ablation_full}); the spectral head's advantage is in the
tail of the coverage curve (Fig.~\ref{fig:cov-cal}a), the regime that governs filter
gating. We do not claim a uniform improvement: in the mid-coverage range the two NLL
heads are comparable, so calibration is a secondary property here, while the primary
covariance result is the tensor transformation law.

\paragraph{Filter-connected covariance stress test.}
To isolate the covariance parameterization from the learned mean, we also freeze the same mean predictor and vary only the covariance head in the SCEKF-connected TLIO pipeline. Nominal ATE is nearly identical, with the coordinate-diagonal head at 1.072 m and the equivariant spectral-derived marginal at 1.081 m. Under a \(20^\circ\) frame error, however, the equivariant parameterization yields 1.349 m versus 1.826 m for the diagonal head; paired differences are significant for \(\theta\geq15^\circ\). Thus the spectral construction is not uniformly better in the nominal setting, but it provides more reliable axis marginals as the sensor/preprocessing frame is perturbed.

The per-window NLL decomposition (Fig.~\ref{fig:cov-cal}b) explains the difference between
the two NLL-trained heads. The diagonal head occupies a narrow range of log-determinants,
whereas the spectral head spans a wide range, indicating that it learns input-dependent
(heteroscedastic) uncertainty rather than a near-constant scale.

\section{Test-Time Frame-Estimation Mismatch: Full Sweeps}
\label{app:calibration_mismatch}

The main text summarizes frame-estimation mismatch with
Figure~\ref{fig:calib_mismatch_full}; here we provide the full numeric table. The
injected angular error is
\begin{equation}
\theta_{\rm err}\in\{3^\circ,5^\circ,10^\circ,15^\circ,20^\circ\}.
\end{equation}
Gravity-frame perturbations corrupt the attitude frame used by gravity-stabilized
pipelines. LFA perturbations corrupt the relative rotations used to express each window
in the final sensor frame. EqNIO O(2) appears only in the gravity-frame columns because
it is not an LFA-based method.

\begin{table*}[htbp]
\centering
\small
\caption{Network-only ATE [m] under test-time frame-estimation mismatch. Lower is better. Gravity columns include EqNIO-style baselines; LFA columns include only methods evaluated with LFA-compatible preprocessing.}
\label{tab:app_calib_mismatch_full}
\renewcommand{\arraystretch}{1.05}
\resizebox{\textwidth}{!}{%
\begin{tabular}{ll|ccccc|ccccc}
\toprule
& & \multicolumn{5}{c|}{Gravity-frame error \(\theta_{\rm err}\)} & \multicolumn{5}{c}{LFA relative-frame error \(\theta_{\rm err}\)} \\
Dataset & Method & \(3^\circ\) & \(5^\circ\) & \(10^\circ\) & \(15^\circ\) & \(20^\circ\)
& \(3^\circ\) & \(5^\circ\) & \(10^\circ\) & \(15^\circ\) & \(20^\circ\) \\
\midrule
\multirow{5}{*}{TLIO}
& ResNet & 1.85 & 1.85 & 2.05 & 2.80 & 4.49 & 2.40 & 2.50 & 3.20 & 4.50 & 5.62 \\
& ResNet+\(\mathrm{SO}(3)\) aug. & 1.97 & 1.97 & 2.00 & 2.05 & 2.10 & 2.55 & 2.65 & 2.85 & 3.15 & 3.55 \\
& EqNIO O(2) & 2.20 & 2.30 & 2.55 & 3.10 & 3.95 & -- & -- & -- & -- & -- \\
& EqNIO O(2)+\(\mathrm{SO}(3)\) aug. & 2.15 & 2.20 & 2.30 & 2.50 & 2.75 & -- & -- & -- & -- & -- \\
& \textbf{GINIO-R} & \textbf{1.85} & \textbf{1.85} & \textbf{1.86} & \textbf{1.88} & \textbf{1.88}
& \textbf{2.32} & \textbf{2.35} & \textbf{2.55} & \textbf{2.60} & \textbf{2.85} \\
\midrule
\multirow{5}{*}{NanoBench}
& ResNet & 0.66 & 0.66 & 0.66 & 0.67 & 0.68 & 0.64 & 0.64 & 0.65 & 0.67 & 0.70 \\
& ResNet+\(\mathrm{SO}(3)\) aug. & 1.20 & 1.21 & 1.22 & 1.24 & 1.26 & 1.21 & 1.22 & 1.23 & 1.23 & 1.27 \\
& EqNIO O(2) & 0.93 & 0.96 & 1.02 & 1.10 & 1.18 & -- & -- & -- & -- & -- \\
& EqNIO O(2)+\(\mathrm{SO}(3)\) aug. & 0.96 & 0.97 & 0.99 & 1.02 & 1.06 & -- & -- & -- & -- & -- \\
& \textbf{GINIO-R} & \textbf{0.56} & \textbf{0.58} & \textbf{0.58} & \textbf{0.59} & \textbf{0.59}
& \textbf{0.55} & \textbf{0.57} & \textbf{0.58} & \textbf{0.59} & \textbf{0.60} \\
\midrule
\multirow{5}{*}{RIDI}
& ResNet & 1.30 & 1.35 & 1.55 & 1.95 & 2.78 & 1.38 & 1.45 & 1.60 & 1.92 & 2.30 \\
& ResNet+\(\mathrm{SO}(3)\) aug. & 1.40 & 1.40 & 1.45 & 1.55 & 1.72 & 1.50 & 1.55 & 1.55 & 1.65 & 1.75 \\
& EqNIO O(2) & 1.45 & 1.50 & 1.65 & 1.90 & 2.35 & -- & -- & -- & -- & -- \\
& EqNIO O(2)+\(\mathrm{SO}(3)\) aug. & 1.42 & 1.45 & 1.50 & 1.60 & 1.75 & -- & -- & -- & -- & -- \\
& \textbf{GINIO-R} & \textbf{1.00} & \textbf{1.05} & \textbf{1.10} & \textbf{1.13} & \textbf{1.22}
& \textbf{1.25} & \textbf{1.28} & \textbf{1.32} & \textbf{1.40} & \textbf{1.50} \\
\bottomrule
\end{tabular}%
}
\end{table*}

The table follows the trend shown in the main text. GINIO-R is the most stable curve on TLIO and RIDI and stays near the noise floor on NanoBench.

\section{Additional Nominal and Filter-Connected Diagnostics}
\label{app:additional_diagnostics}

The aggregate metrics in the main text are complemented here by distribution-level nominal results and representative filter-connected trajectories. These diagnostics are intended to contextualize the headline results rather than establish separate claims: the nominal CDFs show how per-sequence errors are distributed, while the trajectories illustrate representative behavior under a test-time frame rotation with the filter backend held fixed.

\begin{figure*}[t]
\centering
\includegraphics[width=0.94\textwidth]{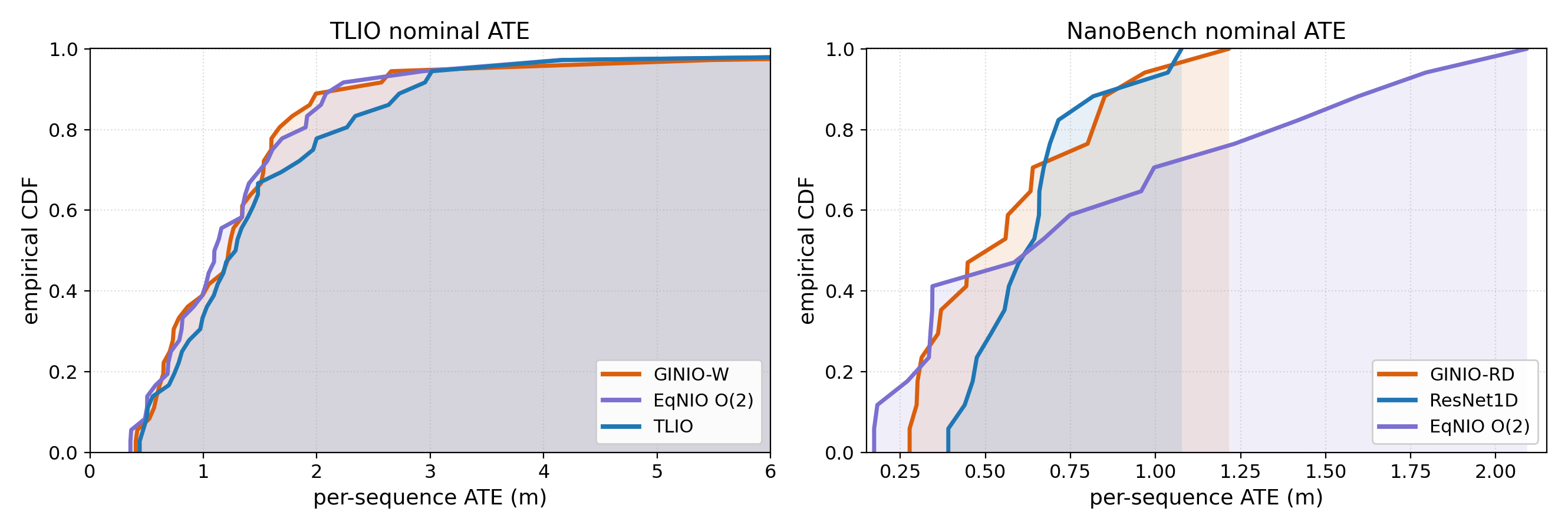}
\caption{Empirical CDFs of per-sequence nominal ATE on TLIO (left) and NanoBench (right). On TLIO, GINIO-W is broadly comparable with the gravity-stabilized baselines across the nominal error distribution; on NanoBench, GINIO-RD shifts the distribution toward lower ATE relative to EqNIO O(2) and is competitive with ResNet1D. These plots support the paper's focus on rotational robustness with competitive nominal performance, rather than a claim of uniform nominal superiority on every sequence.}
\label{fig:nominal_cdfs}
\end{figure*}

Figure~\ref{fig:nominal_cdfs} supplements the mean ATE values in the main tables with per-sequence distributions. In particular, the TLIO curves overlap substantially in the nominal regime even though the same methods separate sharply under rotated-frame evaluation, while the NanoBench distribution shows that the favorable mean ATE of GINIO-RD is not driven by a single sequence.

\begin{figure*}[t]
\centering
\includegraphics[width=0.97\textwidth]{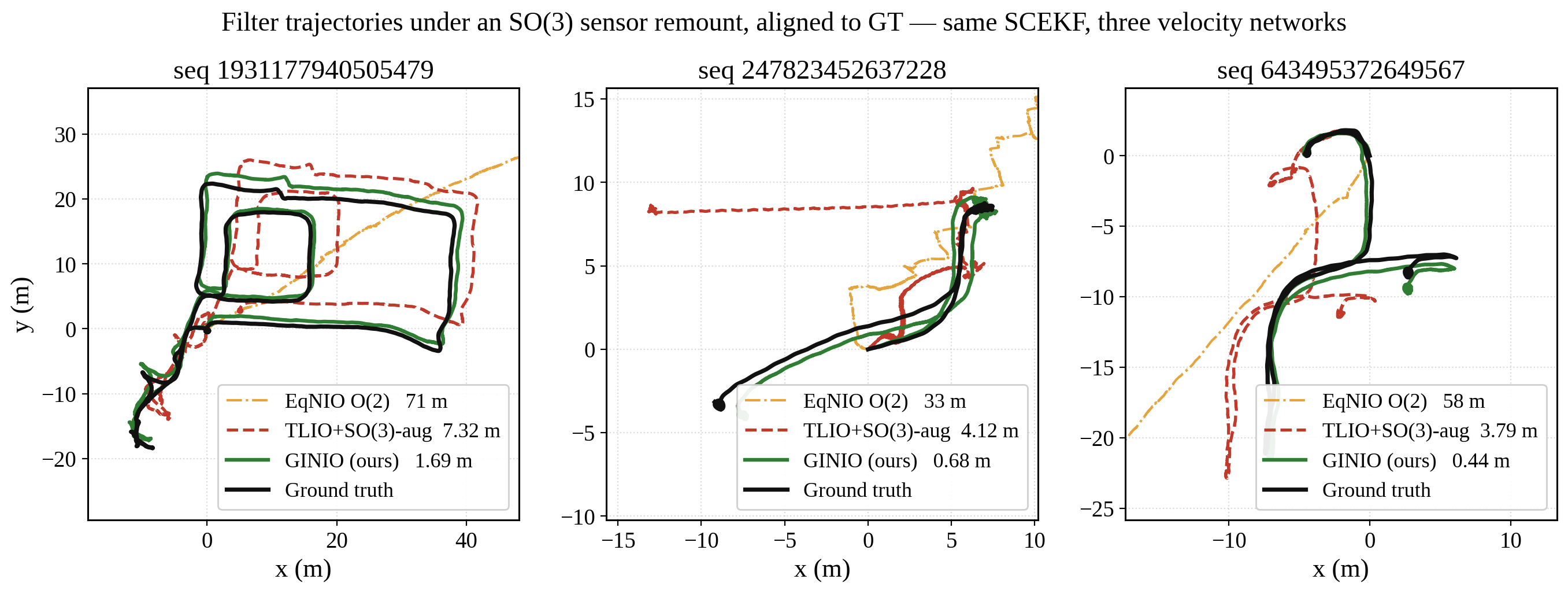}
\caption{Representative filter-connected trajectories under a test-time \(\mathrm{SO}(3)\) measurement-frame rotation, using the same SCEKF backend for all shown methods. GINIO remains close to ground truth in these examples, while the gravity/alignment-dependent baselines can deviate substantially. The three sequences are qualitative examples that complement, but do not replace, the aggregate ATE sweeps in Fig.~\ref{fig:filter_connected_sweeps}.}
\label{fig:filter_rotation_trajectories}
\end{figure*}

The examples in Fig.~\ref{fig:filter_rotation_trajectories} visualize the failure mode summarized by the aggregate filter-connected sweep: when the sensor/preprocessing frame is perturbed, an alignment-dependent learned update can leave its nominal input distribution and corrupt the subsequent filter correction, whereas the equivariant interface keeps the rotated window on the learned transformation orbit. Because only three representative trajectories are shown, we use them as qualitative diagnostics and rely on the full ATE sweeps for the quantitative robustness claim.

\section{Fetch Hardware Validation Details}
\label{app:hardware}

\subsection{Platform and Protocol}

We mounted an Adafruit BNO055 IMU inside the mobile base of a Fetch robot. Only the raw
6-axis inertial channels were used: 3-axis accelerometer and 3-axis gyroscope. BNO055
onboard orientation, Euler angles, gravity vector, linear-acceleration estimate, and
magnetometer outputs were discarded.

All models were trained on 23 nominal-mount sequences only. At test time, the IMU was
physically reinstalled in two unseen \(90^\circ\) orientations---TC1 and TC2, corresponding
to remountings about the \(x\)- and \(y\)-axes (\(\mathbf{g}=-\hat x\) and
\(\mathbf{g}=+\hat y\), versus \(\mathbf{g}=+\hat z\) for the nominal mounting)---and motions
were driven manually and collected independently per mounting. Because the recordings are
not paired across mounting conditions, nominal-to-remount ATE differences are not paired
per-trajectory comparisons; the comparison is therefore at the level of robustness trends.
Within each mounting condition, all architectures are evaluated on the same recorded sequences. Onboard wheel odometry is used as the reference trajectory.

\subsection{Full Hardware Results}

\begin{table}[htbp]
\centering
\small
\caption{Full Fetch remounting results. ATE/RMSE [m], mean \(\pm\) std.}
\label{tab:app_fetch_full}
\setlength{\tabcolsep}{5pt}
\begin{tabular}{llcc}
\toprule
Architecture & Condition & ATE [m] & RMSE [m] \\
\midrule
\multirow{6}{*}{ResNet}
& Normal & 0.105 \(\pm\) 0.082 & 0.118 \(\pm\) 0.091 \\
& Normal+ eval \(\mathrm{SO}(3)\) & 6.234 \(\pm\) 1.483 & 7.208 \(\pm\) 1.739 \\
& TC1 & 7.798 \(\pm\) 5.973 & 9.075 \(\pm\) 6.625 \\
& TC2 & 8.505 \(\pm\) 4.316 & 9.865 \(\pm\) 4.859 \\
& TC1+ eval \(\mathrm{SO}(3)\) & 6.530 \(\pm\) 3.654 & 7.299 \(\pm\) 3.893 \\
& TC2+ eval \(\mathrm{SO}(3)\) & 6.235 \(\pm\) 3.092 & 7.159 \(\pm\) 3.570 \\
\midrule
\multirow{6}{*}{GINIO-R}
& Normal & \textbf{0.095 \(\pm\) 0.099} & \textbf{0.105 \(\pm\) 0.108} \\
& Normal+eval \(\mathrm{SO}(3)\) & \textbf{0.095 \(\pm\) 0.114} & \textbf{0.105 \(\pm\) 0.124} \\
& TC1 & \textbf{0.226 \(\pm\) 0.115} & \textbf{0.258 \(\pm\) 0.135} \\
& TC2 & \textbf{0.764 \(\pm\) 0.468} & \textbf{0.857 \(\pm\) 0.518} \\
& TC1+eval \(\mathrm{SO}(3)\) & \textbf{0.227 \(\pm\) 0.115} & \textbf{0.258 \(\pm\) 0.136} \\
& TC2+eval \(\mathrm{SO}(3)\) & \textbf{0.764 \(\pm\) 0.468} & \textbf{0.857 \(\pm\) 0.518} \\
\bottomrule
\end{tabular}
\end{table}
Rows marked ``eval SO(3)'' apply additional synthetic rotations to the recorded IMU streams at test time; these are not training-time augmentation, which is evaluated separately in Table~\ref{tab:app_fetch_aug}. GINIO-R is invariant to these synthetic rotations to numerical precision, changing by less than \(10^{-3}\) m in ATE.

\subsection{Hardware Augmentation Ablation}
\label{app:hardware_aug}

\begin{table}[htbp]
\centering
\small
\caption{Fetch remounting with training-time rotation augmentation. Full \(\mathrm{SO}(3)\)
augmentation improves the non-equivariant ResNet (unseen mean \(8.15\rightarrow0.58\,\)m)
at a nominal-accuracy cost. GINIO-R generalizes to the unseen mounts without any rotation
augmentation (unseen mean \(0.495\,\)m), though one orientation (TC2, \(0.764\,\)m) remains
harder than the other (TC1, \(0.226\,\)m). Its robustness comes from the architecture
rather than from augmented training.}
\label{tab:app_fetch_aug}
\begin{tabular*}{\textwidth}{@{\extracolsep{\fill}}lcccccc@{}}
\toprule
Model & Normal & TC1 & TC2 & Unseen mean & Gain TC1 & Gain TC2 \\
\midrule
ResNet & 0.105 & 7.798 & 8.505 & 8.152 & 1.0\(\times\) & 1.0\(\times\) \\
GINIO-R & \textbf{0.095} & \textbf{0.226} & 0.764 & 0.495 & \textbf{34.5\(\times\)} & 11.1\(\times\) \\
ResNet+\(90^\circ\) aug. & 1.751 & 1.477 & 2.792 & 2.135 & 5.3\(\times\) & 3.0\(\times\) \\
GINIO-R+\(90^\circ\) aug. & 0.540 & 0.589 & \textbf{0.567} & 0.578 & 13.2\(\times\) & \textbf{15.0\(\times\)} \\
ResNet+\(\mathrm{SO}(3)\) aug. & 0.542 & 0.580 & 0.575 & 0.578 & 13.4\(\times\) & 14.8\(\times\) \\
GINIO-R+\(\mathrm{SO}(3)\) aug. & 0.108 & 0.265 & 0.721 & \textbf{0.493} & 29.4\(\times\) & 11.8\(\times\) \\
\bottomrule
\end{tabular*}
\end{table}

\begin{figure}[t]
\centering
\includegraphics[width=0.85\textwidth]{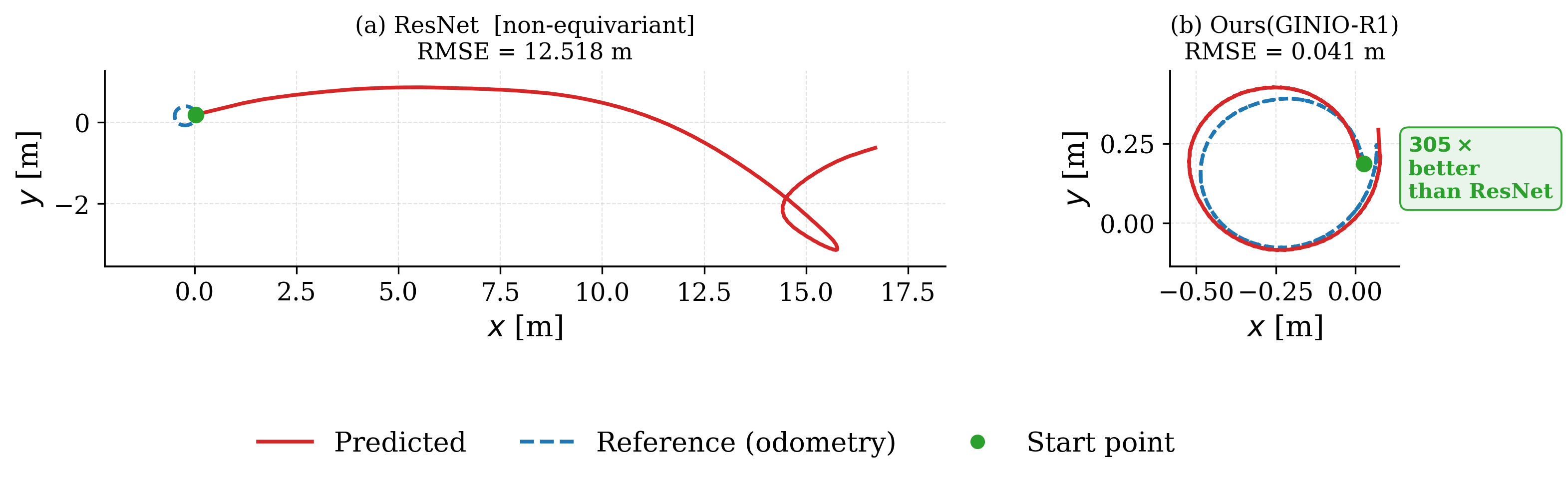}
\caption{Representative Fetch remounting result on a fast-circle sequence under TC1. The non-equivariant ResNet diverges, whereas GINIO-R tracks the onboard odometry reference.}
\label{fig:app_fetch_traj}
\end{figure}

\begin{figure*}[!tbp]
\centering
\begin{subfigure}[t]{\textwidth}
  \centering
  \includegraphics[width=0.9\linewidth,trim=31 34 50 30,clip]{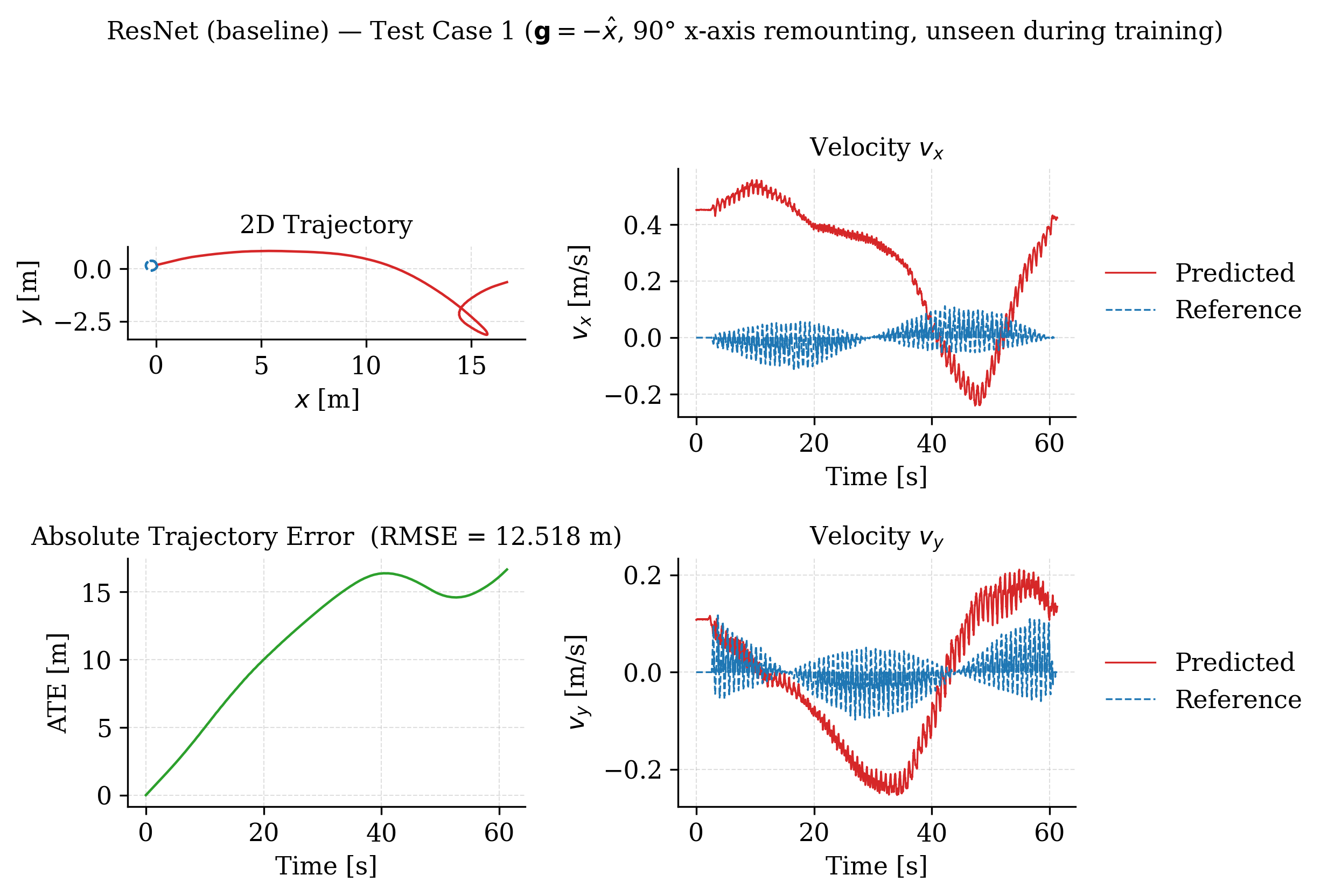}
  \caption{ResNet under TC1.}
  \label{fig:app_fetch_resnet_diag}
\end{subfigure}

\vspace{0.6em}
\begin{subfigure}[t]{\textwidth}
  \centering
  \includegraphics[width=0.9\linewidth,trim=28 34 51 30,clip]{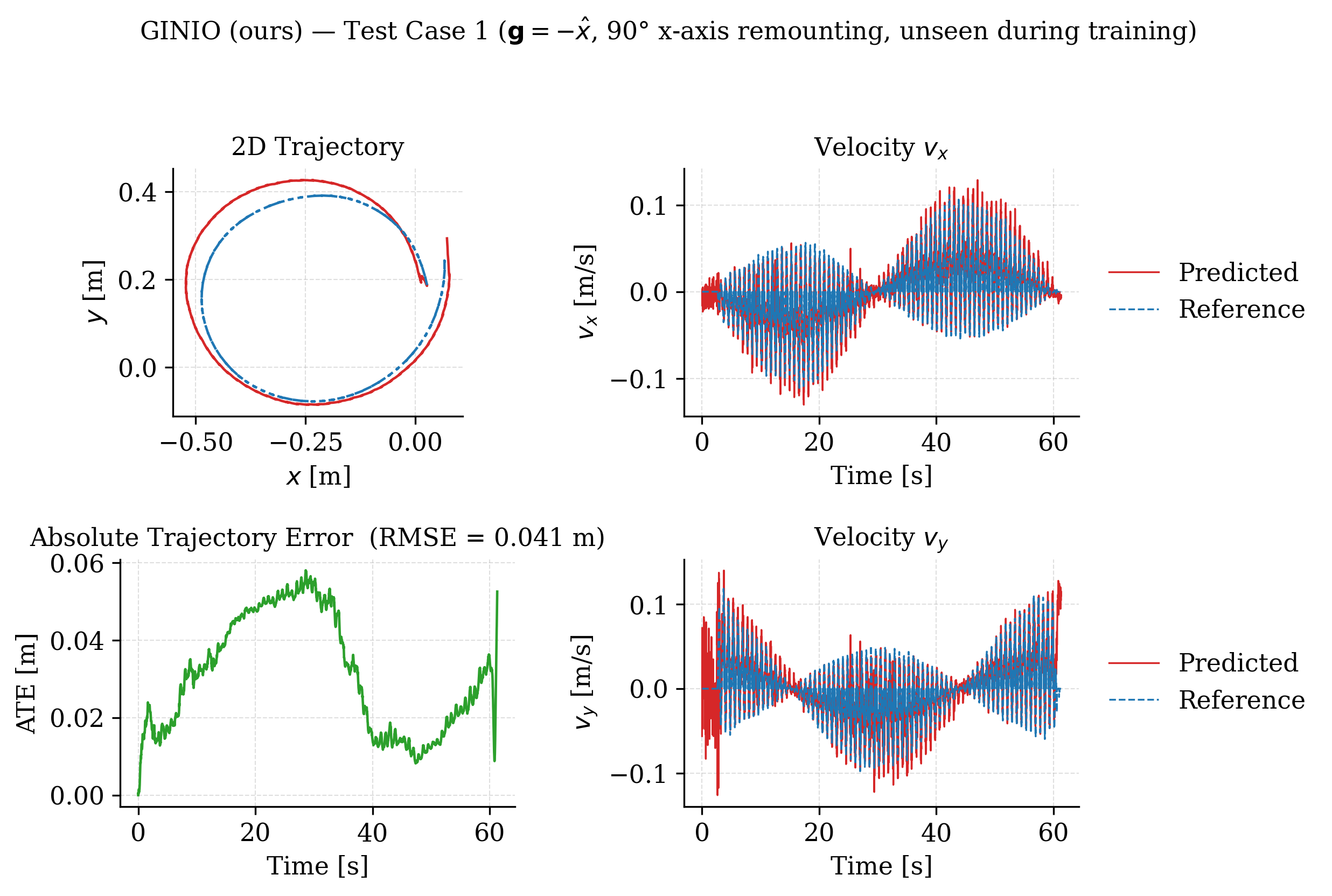}
  \caption{GINIO-R under TC1.}
  \label{fig:app_fetch_ginio_diag}
\end{subfigure}
\caption{Full diagnostic plots under TC1. The non-equivariant ResNet (a) diverges,
whereas GINIO-R (b) tracks the reference.}
\label{fig:app_fetch_diag}
\end{figure*}

\section{Additional Visually Degraded Evaluation}
\label{app:aquatic}

AquaticVision is included as an auxiliary visually degraded evaluation. It is not part of the main headline, which rests on TLIO/RIDI, NanoBench, and Fetch, but it provides an additional setting where vision-based odometry is challenged and IMU-only prediction remains useful.

\begin{table}[htbp]
\centering
\small
\caption{ATE [m] on AquaticVision S04/S06. In the unseen setting, S04 and S06 are excluded
from training. Visual-SLAM baselines have no ID/\(\mathrm{SO}(3)\) entry because they are
not subject to IMU-frame rotation.}
\label{tab:app_aquatic}
\setlength{\tabcolsep}{5pt}
\begin{tabular}{lcccc}
\toprule
\multirow{2}{*}{Method} & \multicolumn{2}{c}{ID/ID} & \multicolumn{2}{c}{ID/\(\mathrm{SO}(3)\)} \\
\cmidrule(lr){2-3}\cmidrule(lr){4-5}
& S04 & S06 & S04 & S06 \\
\midrule
\multicolumn{5}{l}{\emph{Visual SLAM baselines}} \\
VINS-Stereo & 0.0489 & 1.16 & -- & -- \\
ORB-SLAM2 & 0.427 & failed & -- & -- \\
ESVO2 & failed & failed & -- & -- \\
\midrule
\multicolumn{5}{l}{\emph{Inertial odometry, unseen test data}} \\
GINIO & \textbf{0.575} & \textbf{0.473} & \textbf{0.575} & \textbf{0.473} \\
ResNet & 0.583 & 0.693 & 1.010 & 0.884 \\
\midrule
\multicolumn{5}{l}{\emph{Inertial odometry, seen test data}} \\
GINIO & \textbf{0.397} & \textbf{0.347} & \textbf{0.397} & \textbf{0.347} \\
ResNet & 0.431 & 0.693 & 1.120 & 0.774 \\
\bottomrule
\end{tabular}
\end{table}

The AquaticVision result is consistent with the main experiments: the equivariant predictor is invariant under test-time rotations, whereas the non-equivariant baseline degrades under the same perturbation.


\end{document}